\pdfoutput=1

\documentclass[10pt,twocolumn,letterpaper]{article}

\usepackage{iccv}
\usepackage{times}
\usepackage{epsfig}
\usepackage{graphicx}
\usepackage{amsmath}
\usepackage{amsthm}
\usepackage{amssymb}
\usepackage{amsfonts}
\usepackage{booktabs}
\usepackage{mathrsfs}
\usepackage[utf8]{inputenc}
\usepackage[T1]{fontenc}
\usepackage{url}
\usepackage{nicefrac}
\usepackage{microtype}
\usepackage{xcolor}
\usepackage{multirow}
\usepackage{placeins}
\usepackage{bbm}
\usepackage{enumerate}
\usepackage{pgfplots}
\usepackage{cite}
\pgfplotsset{width=10cm,compat=1.9}
\usepgfplotslibrary{groupplots}
\usepackage{hyperref}
\hypersetup{colorlinks,allcolors=black}

\newtheorem{theorem}{Theorem}

\newtheorem{definition}{Definition}

\iccvfinalcopy 

\def\iccvPaperID{****} 

\begin{document}

\title{Marginal Thresholding in Noisy Image Segmentation}

\author{
  Marcus Nordstr\"om \thanks{Author is also affiliated with RaySearch Laboratories.}\\
  Department of Mathematics\\
  KTH Royal Institute of Technology\\
  Stockholm, Sweden \\
  {\tt\small marcno@kth.se} \\
  \and
  Henrik Hult \\
  Department of Mathematics\\
  KTH Royal Institute of Technology\\
  Stockholm, Sweden \\
  {\tt\small{hult@kth.se}} \\
  \and
  Atsuto Maki \\
  Department of Computer Science\\
  KTH Royal Institute of Technology\\
  Stockholm, Sweden \\
  {\tt\small{atsuto@kth.se}} \\
}

\maketitle
\ificcvfinal\thispagestyle{empty}\fi

\begin{abstract}
This work presents a study on label noise in medical image segmentation by considering a noise model based on Gaussian field deformations.
Such noise is of interest because it yields realistic looking segmentations and because it is unbiased in the sense that the expected deformation is the identity mapping.
Efficient methods for sampling and closed form solutions for the marginal probabilities are provided.
Moreover, theoretically optimal solutions to the loss functions cross-entropy and soft-Dice are studied and it is shown how they diverge as the level of noise increases.
Based on recent work on loss function characterization, it is shown that optimal solutions to soft-Dice can be recovered by thresholding solutions to cross-entropy with a particular a priori unknown threshold that efficiently can be computed.
This raises the question whether the decrease in performance seen when using cross-entropy as compared to soft-Dice is caused by using the wrong threshold.
The hypothesis is validated in 5-fold studies on three organ segmentation problems from the TotalSegmentor data set, using 4 different strengths of noise.
The results show that changing the threshold leads the performance of cross-entropy to go from systematically worse than soft-Dice to similar or better results than soft-Dice.
\end{abstract}

\section{Introduction}

In medical image analysis it is often of interest to partition an image into foreground and background with respect to some region of interest.
This problem is referred to as the medical image segmentation problem.
Since the introduction of the U-net~\cite{ronneberger2015u}, state of the art methods for addressing the problem are dominated by neural network methods trained according to the supervised learning paradigm~\cite{du2020medical,siddique2021u}.
These methods are remarkably effective in scenarios with abundant and clean data, and some would argue that the segmentation problem under such circumstances has been solved.
However, manually delineating a region of interest accurately and reliably can be a complex and time-consuming process that often requires medical expertise.
This result in high costs associated with obtaining high-quality data sets for training, and consequently, that data used in practice often is imperfect in the sense that the manually provided labels are contaminated by noise~\cite{bridge2016intraobserver,nir2018automatic,armato2011lung,nyholm2018mr}.

A large amount of research has been devoted to medical image segmentation with imperfect data and
one important question that has been addressed is how robust various segmentation methods are to label noise.
Typically, experiments are conducted by first obtaining some data set, then corrupting the training data with various types of noise and finally training different models on the generated data sets.
The results are then evaluated for the different models using some kind of evaluation metric and compared with a baseline model trained without any added noise.
Simple noise models that have been studied include randomly chopping off parts of the structure~\cite{yu2020robustness} and flipping labels or clusters of labels in an independent manner~\cite{heller2018imperfect}.
A more realistic noise model that has been studied is based on random deformations, where deformations are formed by the interpolation of low dimensional Gaussian arrays.
These Gaussian deformation models are unbiased in the sense that the expected deformation is given by the identity deformation.
It has experimentally been shown that machine learning models are more robust to noise that has this properties as compared to noise that does not have this property~\cite{vorontsov2021label}.

How label noise affects the performance of segmentation models depend on the loss function that is used to train the model.
Arguably, the most classical loss function is cross-entropy.
This loss function is known to have a lot of nice theoretical properties and to be stable in practice.
Because of the unbalanced nature of medical image segmentation caused by the foreground typically being significantly smaller than the background, and the widespread use of Dice as the evaluation metric, an alternative loss based on a smoothed version of the Dice metric was later proposed.
This loss is referred to as soft-Dice or the Dice loss, and has been shown to yield better performance than cross-entropy in several experimental settings at the cost of sometimes introducing some stability issues~\cite{nordstrom2020calibrated,bertels2019optimizing}.
As a consequence, to get some improved performance without introducing to much potential stability issues, it is common to use linear combintations of the losses.
This is for instance the type of loss used in the popular NNUnet~\cite{isensee2021nnu}.

This work takes a closer look on a noise model based on Gaussian deformations.
In particular, such a model is formalized and efficient methods for simulation and closed form expression for marginal probabilities are derived.
For the noise model, optimal solutions to cross-entropy and soft-Dice are then studied,
and it is shown that soft-Dice and cross-entropy yield the same class of optimal solutions when the labels are noise-free, but that they diverge when noise is increased.
It is further shown that optimal solutions to soft-Dice can be recovered by thresholding solutions to cross-entropy with a particular a prior unknown but efficiently computable threshold.
This leads to the question whether cross-entropy with the alternative threshold yields the performance of soft-Dice, and it is experimentally shown that this indeed is the case.

\paragraph{Contributions:}
A family of noise models based on Gaussian field deformations is formalized.
An efficient method for sampling noisy segmentations and a closed form expression for the marginal probabilities are derived.
The noise model is used to show that optimal solutions to cross-entropy and soft-Dice diverge as noise increases, but that optimal solutions to soft-Dice can be recovered by thresholding the cross-entropy estimates with a computable threshold.
This is then validated experimentally on three organs from the TotalSegmentor data set~\cite{wasserthal2022totalsegmentator}.

\section{Related work}
Label noise in medical image segmentation data can be a consequence of time constraints imposed on the annotators drawing the segmentations, differing opinions on where the label borders should be or something else.
The fact that label noise often is present in data used for model training is widely known and have been pointed out in for example~\cite{bridge2016intraobserver, nir2018automatic,armato2011lung,nyholm2018mr}.
Many methods have been proposed for handling label noise.
Examples include active learning methods, where experts are iteratively probed for more information~\cite{gu2018reliable} and re-weighting schemes where methods dynamically learn to put different weights on different regions depending on certainty~\cite{mirikharaji2019learning}.
For fairly recent reviews, see~\cite{tajbakhsh2020embracing,karimi2020deep}.
 
The influence label noise has on various loss functions have been studied in several papers.
This include the influence it has on calibration and volume bias for predictions generated with models trained using cross-entropy and soft-Dice~\cite{bertels2021theoretical}, re-calibration methods for improving performance of soft-Dice~\cite{rousseau2021post} and the relationship between calibration and volume bias~\cite{popordanoska2021relationship}.
Other experimental work showing that the expected calibration error is lower for cross-entropy than for soft-Dice include~\cite{mehrtash2020confidence}.
In the above, noise has been analyzed by considering the expected score with respect to some noisy segmentation.
An alternative way to incorporate noise is to make use of soft labels.
Examples of work using soft labels with soft-Dice include~\cite{gros2021softseg,kats2019soft}.
Examples of work on soft labels with other loss functions include~\cite{silva2021using,lemay2022label,li2020superpixel}.

Inspired by threshold classifiers studied in~\cite{zhao2013beyond,lipton2014optimal}, optimal segmentations to Accuracy and Dice under label noise was analyzed in~\cite{nordstrom2022image}.
A characterization of optimal segmentations was provided, sharp bounds with respect to volume bias of the optimal segmentations where shown and the volume of the optimizers associated with the metrics was compared.
A similar analysis was later provided for the loss function soft-Dice~\cite{nordstrom2023noisyimage}.
In particular, a characterization of the optimal solutions and the associated volumes are provided.
Moreover, a calibration result showing that a sequence the converge to optimal soft-Dice, when thresholded appropriately, converges to optimal Dice.

The robustness of medical image segmentation to noise has been addressed in several work.
In~\cite{yu2020robustness}, robustness is studied by randomly chopping off sub-structures in the training data.
Of special importance to this work is segmentation noise provided by Gaussian field deformations.
Two works have studied such noise in an experimental setting.
This includes \cite{heller2018imperfect} and \cite{vorontsov2021label}.
In these works, the segmentations in the training data is perturbed according to a low dimensional Gaussian field interpolated to the pixel space.
Models are then trained for various noise levels and results obtained are reported.
Finally, in~\cite{le2016sampling}, a noise model for target delineation in radiotherapy is discussed.
The model is based on taking the level set of Gaussian field and efficient methods for sampling is provided.

\section{Preliminaries}
\subsection{Notation}
Let $\Omega=[0,1]^n\subset \mathbb{R}^n$ be the unit cube of dimension $n\ge 1$ and $\lambda$ be the  standard Lebesgue measure associated with $\mathbb{R}^n$ such that $\lambda(\Omega)=1$.
Let $\mathcal{S}$ be the space of measurable functions from $\Omega$ to $\{0,1\}$, $\mathcal{M}$ be the space of measurable functions from $\Omega$ to $[0,1]$ and $\mathcal{F}$ be the space of bounded measurable functions from $\Omega$ to  $\mathbb{R}$, where all of the spaces $\mathcal{S}$, $\mathcal{M}$ and $\mathcal{F}$ are equipped with their associated Borel $\sigma$-fields and the $L_p$-norms $\lVert \cdot \rVert_p$. 
For short, the notation $\bar{f}(\omega) = 1-f(\omega),\omega\in\Omega$  will be used for any $f\in\mathcal{F}$.
The letter $s\in\mathcal{S}$ is generally used to denote a segmentation, where $s(\omega) = 1$ indicates foreground of the target structure of interest and $s(\omega) = 0$ indicates background.
The letter $m\in\mathcal{M}$ is generally used to denote a marginal function, that is, $m(\omega)\in[0,1]$ indicates the probability of a noisy label occupying site $\omega\in\Omega$.
The letter $c\in \mathcal{M}$ is generally used to denote the soft segmentation that is obtained by optimizing the chosen loss function prior to any thresholding.
The letter $f\in\mathcal{F}$ is generally used to denote a logit function usually associated with a soft segmentation such that $\sigma \circ f(\omega)=c(\omega)$ for $\omega\in\Omega$ a.e., where $\sigma(x)=1/(1+e^{-x})$ is the standard sigmoid function.
Composition of functions are denoted by $\circ$ and $I_{A}(x)$ is used to denote the indicator function over the set $A$, that is $I_{A}(x) = 1$ if $x\in A$ and $I_{A}(x) = 0$ if $x\not\in A$.
The indicator function is used in conjunction with some element $c\in \mathcal{M}$ to denote thresholding of a soft segmentation.
That is, $I_{[t,1]}\circ c$ assigns $1$ to all $\omega\in\Omega$ where $c(\omega) \ge t$ and $0$ to all $\omega\in\Omega$ where $c(\omega) < t$.

\subsection{Loss functions and evaluation metrics}
The two most popular loss functions in the field of medical image segmentation are cross-entropy and soft-Dice.
When noise is present, the label becomes a random variable and the losses need to be extended to a functional over one deterministic soft segmentation (prediction) and one random segmentation (label).
In this work, the soft labeling convention for this extension is adopted for the theoretical analysis
~\cite{kats2019soft, gros2021softseg,silva2021using,lemay2022label,li2020superpixel}.
Formally, the loss functions are defined with respect to a marginal function $m\in\mathcal{M}$ which corresponds to the mean of a noisy segmentation $L$ taking values in $\mathcal{S}$, that is $m(\omega) = L(\omega), \omega \in \Omega$.
\begin{definition}
For any $m\in \mathcal{M}$, (binary) cross-entropy is given by
\begin{align}
\mathrm{CE}_m(c) \doteq
\begin{split}
     -\int_\Omega [&m(\omega)\log(c(\omega))  + \\
    &\bar{m}(\omega) \log(\bar{c}(\omega))] \lambda(d\omega),
\end{split}
    \quad c\in\mathcal{M}.
    \label{eq:ce}
\end{align}
\end{definition}
\begin{definition}
For any $m\in \mathcal{M}$, soft-Dice is given by
\begin{align}
    \mathrm{SD}_m(c) &\doteq 1-\frac{2\int_\Omega c(\omega)m(\omega) \lambda(d\omega)}{\lVert c\rVert_1 + \lVert m\rVert_1}, \quad c\in\mathcal{M}.\label{eq:sd}
\end{align}
\end{definition}

Arguably, the most classical evaluation metric used is Accuracy.
This metric simply measures the fraction of the domain that is correctly labeled.
The most popular evaluation metric on the other hand is the S\"orensen-Dice coefficient, or Dice for short.
To handle label noise, the metrics are generalized in the same way as the loss functions, that is, by using soft labels.

\begin{definition}
For any $m\in \mathcal{M}$, Accuracy for is given by
\begin{align}
    \mathrm{A}_m(s) &\doteq \int_\Omega [s(\omega)m(\omega)  + \bar{s}(\omega)\bar{m}(\omega)] \lambda(d\omega),
     s\in\mathcal{S} 
    \label{eq:acc}
\end{align}

\end{definition}
\begin{definition}
For any $m\in \mathcal{M}$, Dice is given by
\begin{align}
    \mathrm{D}_m(s) &\doteq \frac{2\int_\Omega s(\omega)m(\omega) \lambda(d\omega)}{\lVert s\rVert_1 + \lVert m\rVert_1}, \quad s\in\mathcal{S}.
    \label{eq:dice}
\end{align}
\end{definition}

When doing the experiments, only one label per image is available and it is more accurate to think of the setup as an empirical approximation of the expected score.
For cross-entropy and Accuracy, the definitions using expected score and the definitions using soft labels are equal, that is $\mathbb{E}[\mathrm{A}_L(s)] = \mathrm{A}_m(s)$ and $\mathbb{E}[\mathrm{CE}_L(s)] = \mathrm{CE}_m(s)$.
For soft-Dice and Dice, the definitions are equal provided that the volume of the noisy label is constant, that is $\text{Var}[\lVert L\rVert_1 ] =0$.
If this holds approximately which is often the case in medical image segmentation, where noise often alters the details of the boundary without significantly altering the volume of the segmentation, then it follows that
$\mathbb{E}[\mathrm{D}_L(s)] \approx \mathrm{D}_m(s)$ and similarly that $\mathbb{E}[\mathrm{SD}_L(s)] \approx \mathrm{SD}_m(s)$.


\subsection{Gaussian fields}
Amongst the most central objects in probability theory and statistics are normally distributed random variables, also referred to as Gaussian random variables.
Such random variables appear all over science and engineering and numerous generalizations have been greatly studied because of their richness and interesting properties.
One simple generalization of the classical Gaussian variable is to the multidimensional case for which one talks about multivariate Gaussian variables.
Such a variable, when centered and isotropic, has the following probability density function
\begin{align}
    p_{\sigma^2}(\omega) \dot=
    \frac{1}{(2\pi\sigma^2)^{n/2}} \exp\left\{ -\frac{\lVert \omega\rVert_2^2}{2\sigma^2}\right\}, \quad \omega \in \mathbb{R}^n, \label{eq:density}
\end{align}
where $\sigma^2$ denotes the variance in each of the dimensions.
A common further generalization that can be done is to replace the index set with a subset of the real line, in which case the resulting object is referred to as a Gaussian process.
Similarly, but less common, is when the index set is replaced by a subset of a real euclidean space. 
In this case the resulting object is referred to as a Gaussian field.
Formally, a Gaussian field $Y(\omega),\omega\in\mathbb{R}^n$ is a random variable
such that for any finite collection of elements $\{\omega_1,\dots,\omega_M\}\subset\Omega$,  it holds that
\begin{align}
    (Y(\omega_1),\dots,Y(\omega_m))\sim \text{normal}(0,\Sigma).
\end{align}
The covariance matrix is usually expressed using a $\emph{kernel function}$
\begin{align}
    \Sigma_{m,m'} = k(\omega_m,\omega_{m'}), \quad m,m' \in \{1,\dots,M\}.
\end{align}
Many types of kernel functions with different properties have been studied in the literature.
Amongst the most common are
\begin{align}
    k_{a,b}(\omega,\omega') = a^2\exp\left( -\frac{\lVert \omega-\omega'\rVert_2^2}{2b^2} \right), \omega,\omega'\in\mathbb{R}^n.
    \label{eq:kernel}
\end{align}
This covariance function will also be central in this work.
An important property of Gaussian processes with such covariance functions are that samples are almost surely continuous.

\section{Theory}

\subsection{Optimal segmentations}
When a label is free from noise, any reasonable loss function or metric is optimized uniquely by the target label.
When however there is noise present, it is no longer clear what the solution should be and the losses and metrics implicitly assigns a set of them.
As a consequence, it is natural to investigate what the optimal solutions are to a chosen loss or metric.
For the case of the cross-entropy loss, it is well known that the optimizers are uniquely given by the marginal function, that is 
\begin{align}
    \mathrm{CE}_m(c) = \inf_{c'\in\mathcal{M}} \mathrm{CE}_m(c') \iff
    c(\omega)=m(\omega).
\end{align}
$\lambda$-a.e.
In other words, minimizing cross-entropy yields estimates of the marginal probability distribution associated with the label.
For soft-Dice it was shown in~\cite{nordstrom2023noisyimage} that
\begin{align}
\begin{split}
    &\mathrm{SD}(c) = \inf_{c'\in\mathcal{M}} \mathrm{SD}(c') \iff \\
    &c(\omega)\in
    \begin{cases}
        \{1\} &\text{ if } m(\omega) > \sup_{s'\in\mathcal{S}}\mathrm{D}_m(s')/2, \\
        [0,1] &\text{ if } m(\omega) = \sup_{s'\in\mathcal{S}}\mathrm{D}_m(s')/2, \\
        \{0\} &\text{ if } m(\omega) < \sup_{s'\in\mathcal{S}}\mathrm{D}_m(s')/2, \\
    \end{cases}
    \end{split}
\end{align}
$\lambda$-a.e.
Using the setup adopted in this paper, the optimal segmentations for Accuracy and Dice was characterized in~\cite{nordstrom2022image}.
For Accuracy, the solutions are given by
\begin{align}
\begin{split}
    &\mathrm{A}_m(s) = \sup_{s'\in\mathcal{S}} \mathrm{A}_m(s') \iff \\
    &s(\omega) \in 
    \begin{cases}
        \{1 \} &\text{ if } m(\omega)> 1/2, \\
        \{0,1\} &\text{ if } m(\omega) = 1/2, \\
        \{0\} &\text{ if } m(\omega) <1/2, \\
    \end{cases}
    \end{split}
    \label{eq:char_acc}
\end{align}
$\lambda$-a.e.
Similarly, for Dice the optimal segmentations are given by
\begin{align}
\begin{split}
    &\mathrm{D}_m(s) = \sup_{s'\in\mathcal{S}} \mathrm{D}_m(s') \iff \\
    &s(\omega) \in 
    \begin{cases}
        \{1 \} &\text{ if } m(\omega) > \sup_{s'\in\mathcal{S}}\mathrm{D}_m(s')/2, \\
        \{0,1\} &\text{ if } m(\omega) = \sup_{s'\in\mathcal{S}}\mathrm{D}_m(s')/2, \\
        \{0\} &\text{ if } m(\omega) < \sup_{s'\in\mathcal{S}}\mathrm{D}_m(s')/2, \\
    \end{cases}
    \end{split}
    \label{eq:char_dice}
\end{align}
$\lambda$-a.e.
Ideally, the minimizing of a loss function leads to the maximization of the desired target metric.
That is, that the soft segmentations, when thresholded with a $1/2$-threshold, provides segmentations that maximize the target metric.
Using these characterizations, the $1/2$-thresholded optimizers to the losses can easily be connected to the optimizers of the metrics.
For cross-entropy it follows that
\begin{align}
\begin{split}
    &\mathrm{CE}_m(c) = \inf_{c'\in\mathcal{M}}\mathrm{CE}(c') \implies \\
    &\mathrm{A}_m(I_{[1/2,1]}\circ c) = \sup_{s'\in\mathcal{S}}\mathrm{A}(s').
\end{split}
\label{eq:ce_cal}
\end{align}
Similarly, for soft-Dice it follows that
\begin{align}
\begin{split}
    &\mathrm{SD}_m(c) = \inf_{c'\in\mathcal{M}}\mathrm{SD}(c') \implies \\
    &\mathrm{D}_m(I_{[1/2,1]}\circ c) = \sup_{s'\in\mathcal{S}}\mathrm{D}(s').
\end{split}
 \label{eq:sd_cal}
\end{align}

Based on the characterization of the optimizers to soft-Dice and cross-entropy, it is clear that that an optimizer to soft-Dice can be obtained by first optimizing cross-entropy and then picking a particular threshold that is not known a priori
\begin{align}
    t_m =  \sup_{s'\in\mathcal{S}} \mathrm{D}_m(s')/2~\label{eq:threshold}.
\end{align}
It then then follows that
\begin{align}
\begin{split}
    &\mathrm{CE}_m(c) = \inf_{c'\in\mathcal{M}}\mathrm{CE}(c') \implies \\
    &\mathrm{SD}_m(I_{[t_c,1]}\circ c) = \inf_{c'\in\mathcal{M}}\mathrm{SD}(c').
    \end{split}
    \label{eq:cal_thresh}
\end{align}
This method is a variation of the method presented in~\cite{lipton2014optimal} from the statistical learning literature.
In the medical image segmentation community, however, it has to the best of our knowledge not been treated.


A natural question to address after making these observations is whether the difference in performance associated with cross-entropy and soft-Dice can be related to what threshold is used for cross-entropy.
Such an hypothesis is further strengthened by two papers that have addressed the question of optimal thresholds~\cite{bice2021sensitivity,hussein2022study}.
The results from these papers experimentally show that using $1/2$ as a common threshold leads to sub-optimal segmentations for various loss functions with respect to Dice.
However, neither is a method for computing thresholds uniquely for every image considered, nor is the connection to the theoretical optimal threshold done~\eqref{eq:threshold}.

In~\cite{nordstrom2020calibrated}, it was proposed that the instability sometimes observed for soft-Dice may be associated with the fact that it is not convex (nor quasi-convex) when using sigmoid functions to enforce the bounds, that is $f\mapsto \mathrm{SD}_m(\sigma\circ f), f\in\mathcal{F}$.
This is in contrast to cross-entropy where the corresponding $f\mapsto\mathrm{CE}_m(\sigma\circ f), f\in\mathcal{F}$ is known to be convex.
If cross-entropy with the alternative threshold yields the same performance as soft-Dice, then using it may provide the performance of soft-Dice without sacrificing the stability of cross-entropy.

\subsection{Label noise}
In contrast to binary classification, where a lot of label noise can be modeled by simple independent flipping of binary variables, label noise in segmentation can in general have very complicated dependency structures.
Often labels affected by noise have the appearance of being drawn by someone with \emph{shaky hands}.
For such noise, the correlation between border points will depend on the distance between them.
That is, if two points are close to each other, then the correlation will be high and vice versa.

One way to model segmentations that look as if they have been drawn by shaky hands is by means of random Gaussian field deformations.
In particular, let
\begin{align}
 X=(X_1,\dots,X_n)^T \label{eq:X}
\end{align}
 be such that $\{X_i\}_{i=1}^n$ are independent Gaussian fields with kernel function~\eqref{eq:kernel}.
This means that $X$ is a random vector field taking values in $\mathbb{R}^n\mapsto\mathbb{R}^n$.
With such a random vector field, a random segmentation is formed by deforming a noise-free segmentation $l\in \mathcal{S}$ as follows
\begin{align}
L(\omega) = 
\begin{cases}
    l(\omega+X(\omega)) &\text{ if } \omega + X(\omega) \in \Omega, \\
    0 & \text{ if } \omega + X(\omega) \not\in \Omega.
\end{cases}, \omega\in\Omega.
\label{eq:noisysegmentation}
\end{align}

How samples of the random segmentations look depend on what parameters are chosen when specifying the kernel function.
The parameter $a$ controls how large the deformations are, where greater $a$ leads to larger deformations and vice versa.
The parameter $b$ controls how smooth the deformations are, where greater $b$ leads to smoother deformations and vice versa.
Modeling noisy segmentations with~\eqref{eq:noisysegmentation} is convenient because it allows for the derivation of several theoretical properties.
In this work, three such important properties are proved.

In Theorem~\ref{theorem:marginal}, it is shown that the marginal probabilities associated with a noisy segmentation can be computed by convolving the associated noise-free segmentation label with an appropriate Gaussian density.
\begin{theorem}
Let $l$ and $L$ be defined as in~\eqref{eq:noisysegmentation} and $p_{\sigma^2}$ be defined as in~\eqref{eq:density}, then if follows that
\begin{align}
    \mathbb{E}[L(\omega)] = \int_\Omega l(\omega')p_{a^2}(\omega-\omega') \lambda(d\omega'), \quad \omega\in\Omega.
\end{align}
\label{theorem:marginal}
\end{theorem}
In Theorem~\ref{theorem:volume}, the marginal representation is used to show that the the volume of the noisy segmentation is approximately unbiased in the sense that the expected volume of a noisy segmentation is approximately equal to the volume of the corresponding noise-free segmentation.
The reason why this does not hold exactly is that there is some probability for the target structure to deform in such a way that parts of the structure leave the domain $\Omega$.
For most reasonable choices of parameters, the probability of this happening is negligible.
\begin{theorem}
Let $l$ and $L$ be defined as in~\eqref{eq:noisysegmentation} and $p_{\sigma^2}$ be defined as in~\ref{eq:density}, then it follows that
\begin{align}
\mathbb{E}[\lVert L\rVert_1] =  \lVert l \rVert_1 - \xi,
\end{align}
where 
\begin{align}
    \xi = \int_{\Omega}\int_{\mathbb{R}^n\setminus \Omega} l(\omega') p_{a^2}(\omega-\omega') \lambda(d\omega)\lambda(d\omega').
\end{align}
\label{theorem:volume}
\end{theorem}

In Theorem~\ref{theorem:representation}, it is shown that $X$~\eqref{eq:X} is equal in distribution to a particular vector of stochastic integrals.
In the case $n=1$, this could have been done using regular Ito-integrals, but for the case $n>1$ a multivariate counter part is needed.
For this scattered Gaussian measures are used.
\begin{theorem}
Let $X$ be as in~\eqref{eq:X}, $p_{a^2}$ be as in~\eqref{eq:density} and $W_1,\dots,W_n$ be independent copies of independently scattered Gaussian measures on $\mathbb{R}^n$ with control measure $\lambda$.
Then it follows that
\begin{align}
    X(\omega) \overset{d}{=}  
    \begin{pmatrix}
    a (2\pi b^2)^{n/4} \int_{\mathbb{R}^n} p_{b^2/2}(\omega-\omega') W_1(dw') \\
    \vdots \\
    a (2\pi b^2)^{n/4} \int_{\mathbb{R}^n} p_{b^2/2}(\omega-\omega') W_n(dw')
    \end{pmatrix}.
\end{align}
\label{theorem:representation}
\end{theorem}

Details on scattered Gaussian measures and the proof can be found in the Supplementary Document.
The important consequence of this theorem is that when the domains are discretized, it implies that noisy segmentations can be sampled by drawing independent Gaussian distributed variables associated with all of the voxels, and then smoothing the associated random arrays with Gaussian filters.
Both operations which are very effectively implemented.
This allows for efficient sampling of noisy segmentations, even when the image is composed of million of voxels.


\begin{figure*}[htb!]
  \centering
\begin{tikzpicture}
  \begin{groupplot}[group style={group size=7 by 4, horizontal sep=0.05cm, vertical sep=0.05cm},height=3.9cm,width=3.9cm,xmin=0,xmax=1,ymin=0,ymax=1, xticklabels=\empty, yticklabels=\empty]
  
    \nextgroupplot[title={$a=0.00$}, ylabel={Samples}]
    \addplot graphics[xmin=0,xmax=1,ymin=0,ymax=1] {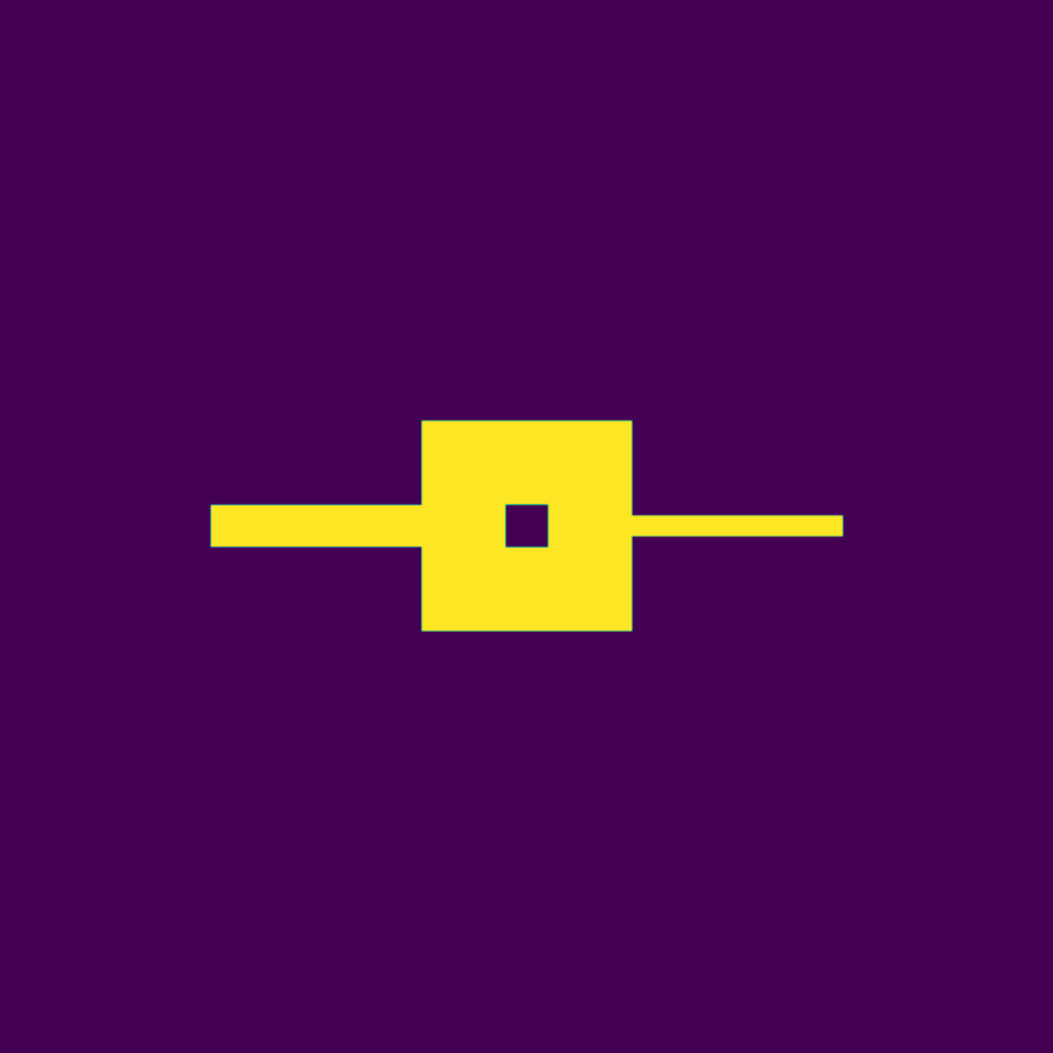};
    
    \nextgroupplot[title={$a=0.01$}]
    \addplot graphics[xmin=0,xmax=1,ymin=0,ymax=1] {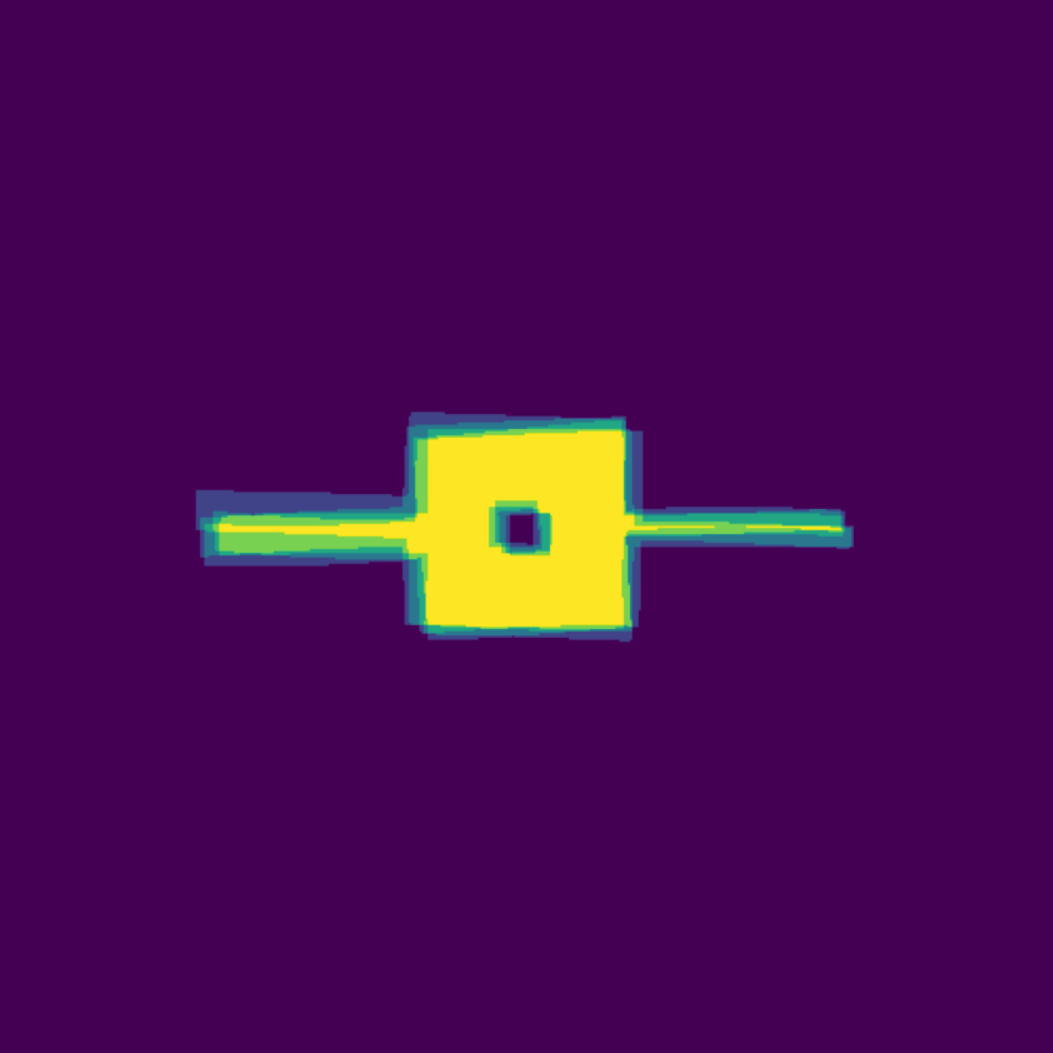};
    
    \nextgroupplot[title={$a=0.02$}]
    \addplot graphics[xmin=0,xmax=1,ymin=0,ymax=1] {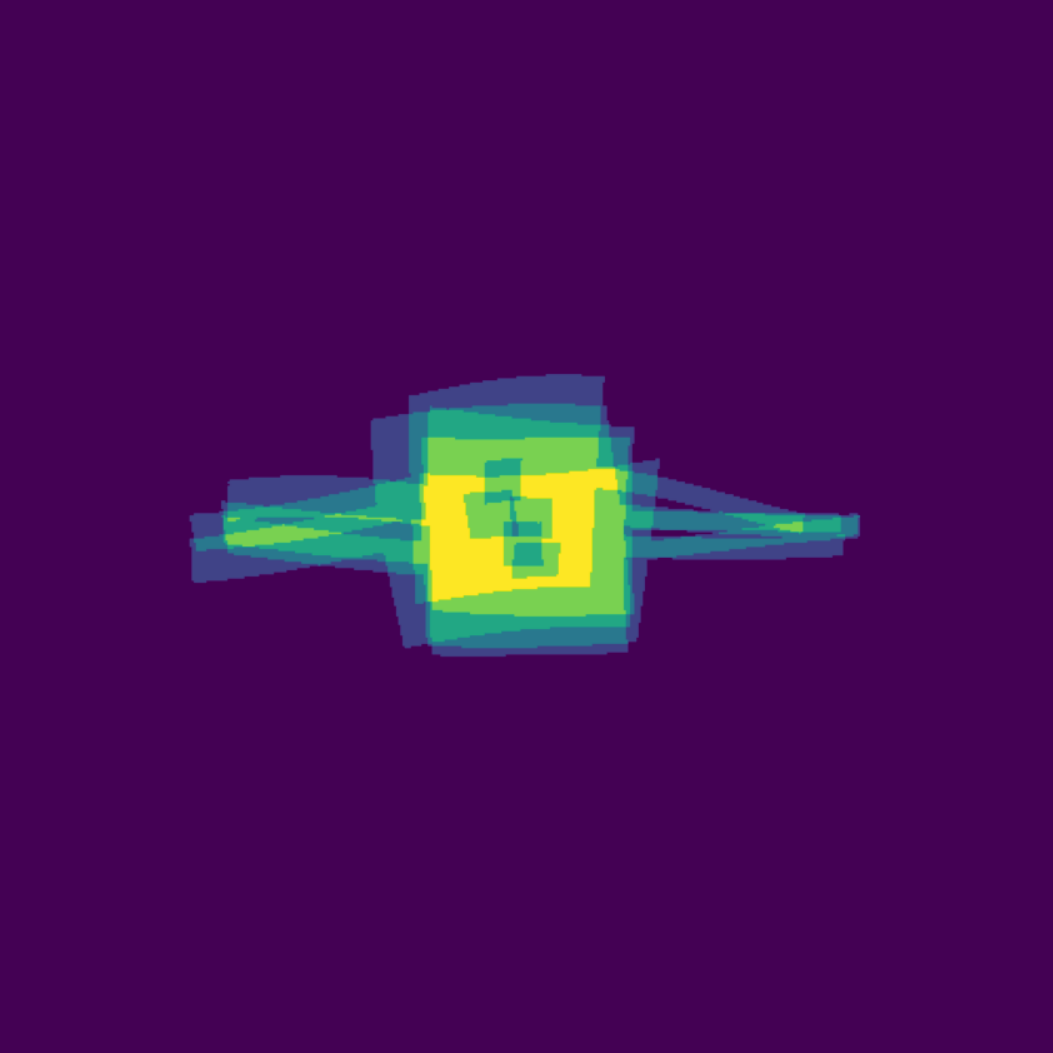};
  
    \nextgroupplot[title={$a=0.03$}]
    \addplot graphics[xmin=0,xmax=1,ymin=0,ymax=1] {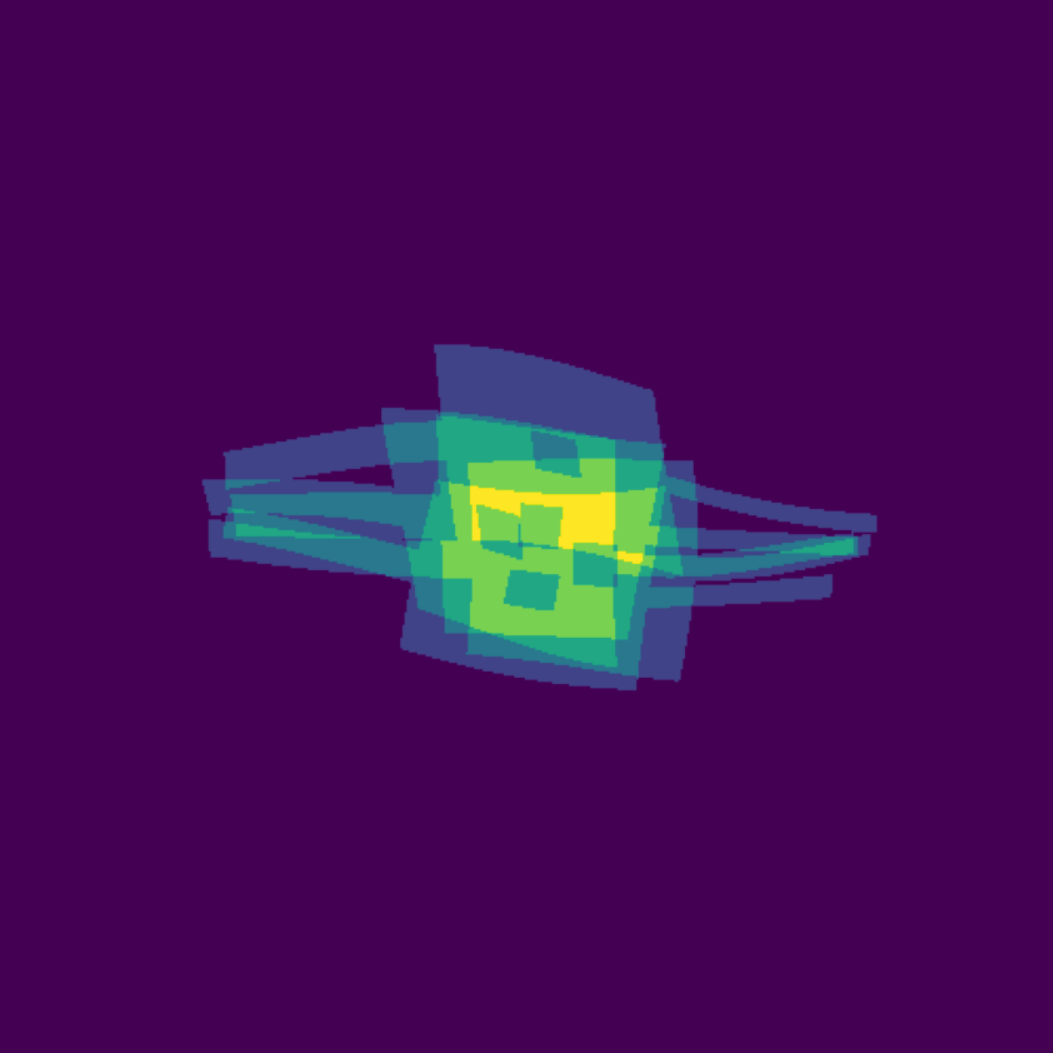};
 
    \nextgroupplot[title={$a=0.04$}]
    \addplot graphics[xmin=0,xmax=1,ymin=0,ymax=1] {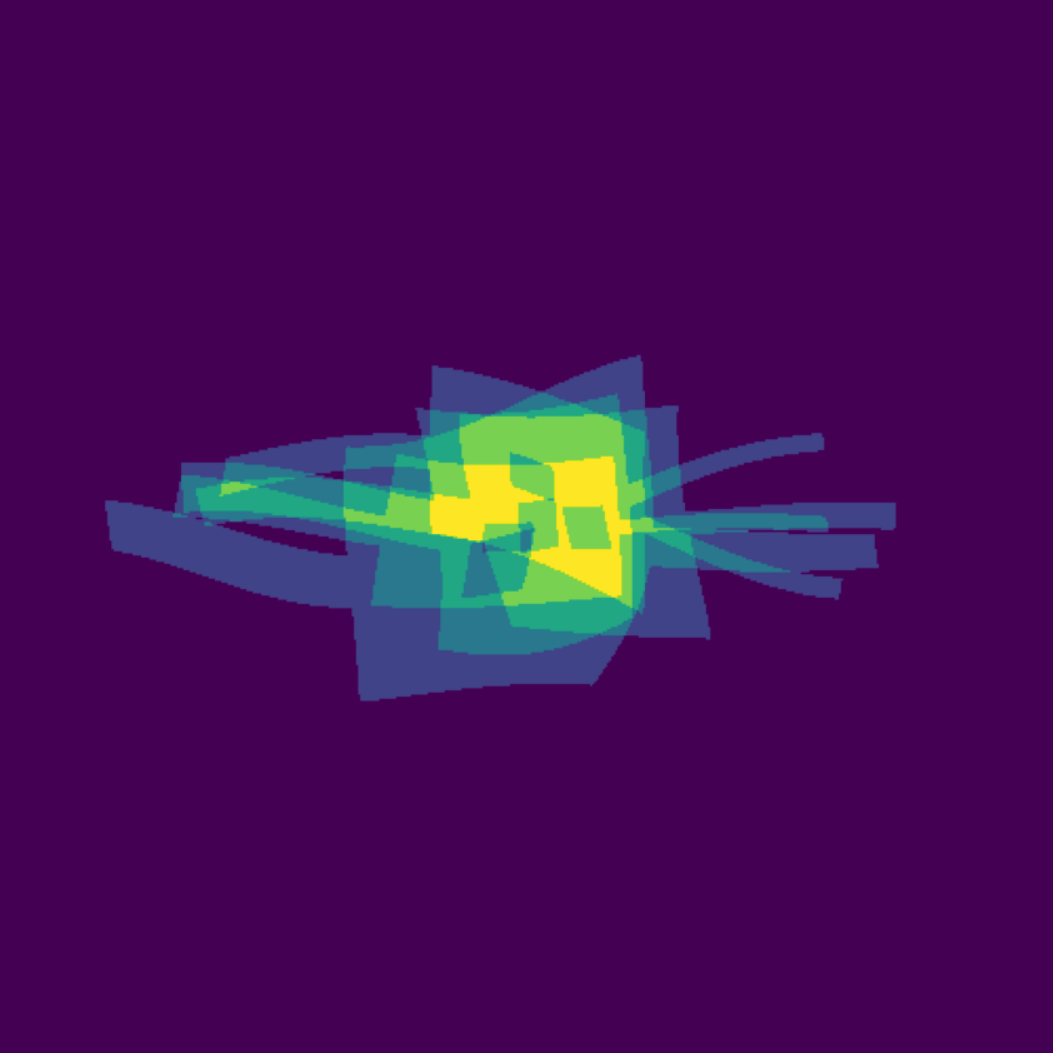};

    \nextgroupplot[title={$a=0.05$}]
    \addplot graphics[xmin=0,xmax=1,ymin=0,ymax=1] {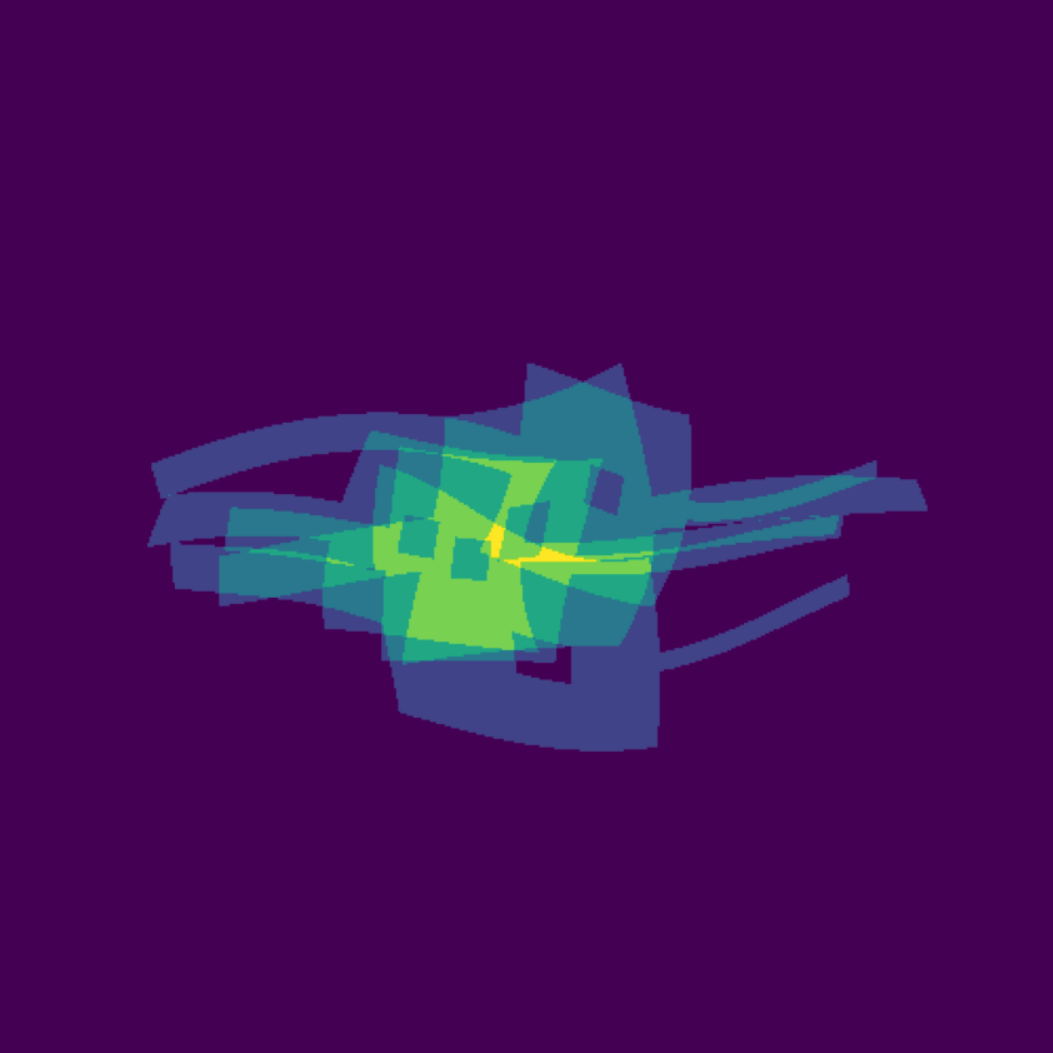};
 
    \nextgroupplot[title={$a=0.06$}]
    \addplot graphics[xmin=0,xmax=1,ymin=0,ymax=1] {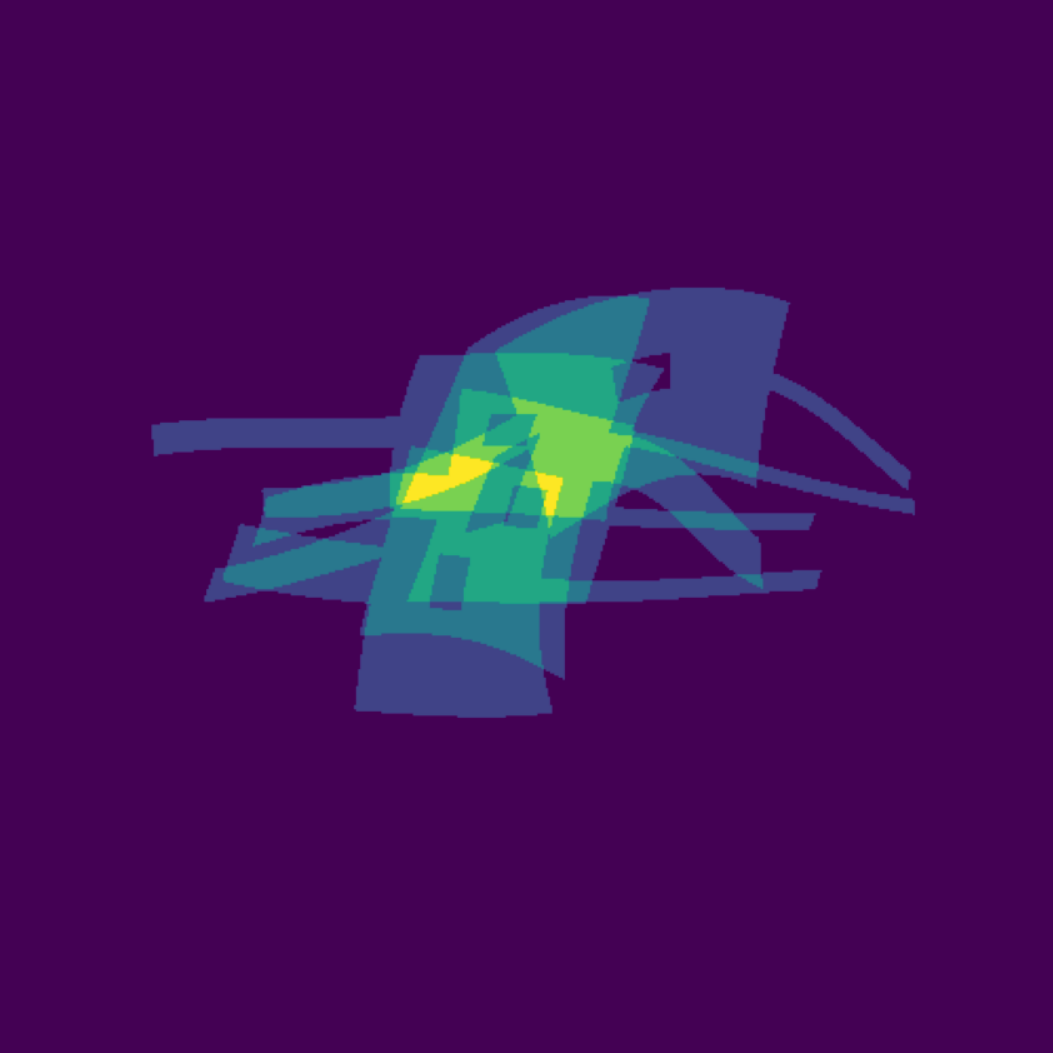};
    
    
    \nextgroupplot[ylabel={Marginal} ]
    \addplot graphics[xmin=0,xmax=1,ymin=0,ymax=1] {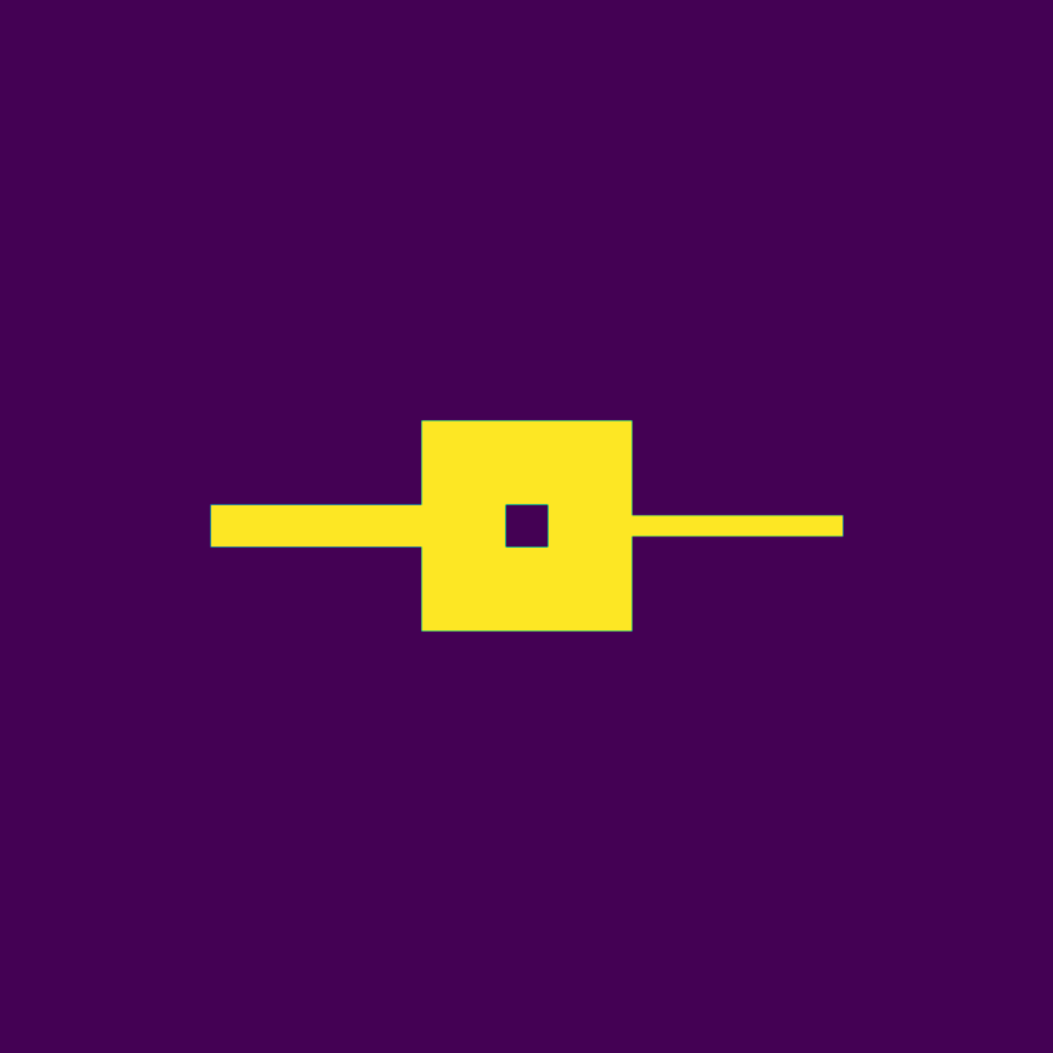};
    
    \nextgroupplot
    \addplot graphics[xmin=0,xmax=1,ymin=0,ymax=1] {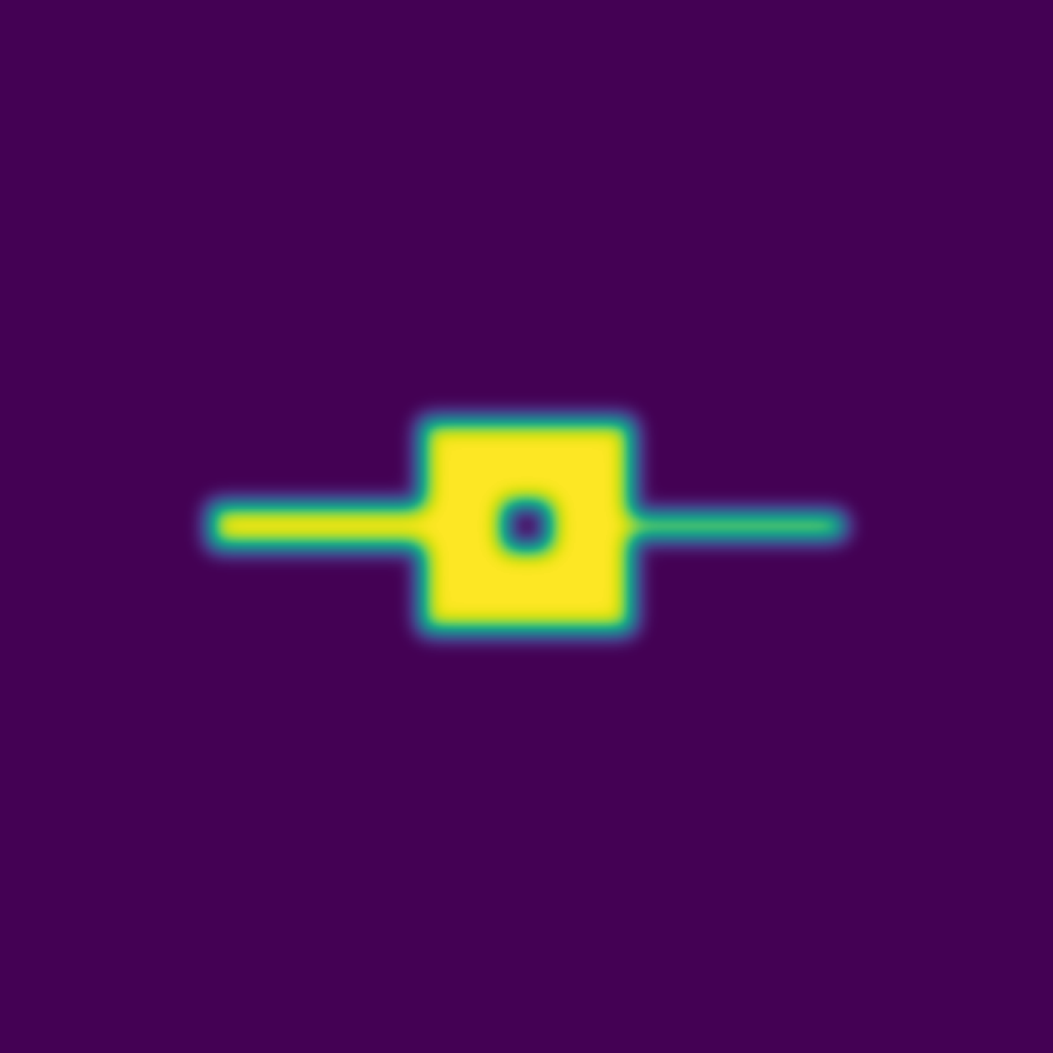};
    
    \nextgroupplot
    \addplot graphics[xmin=0,xmax=1,ymin=0,ymax=1] {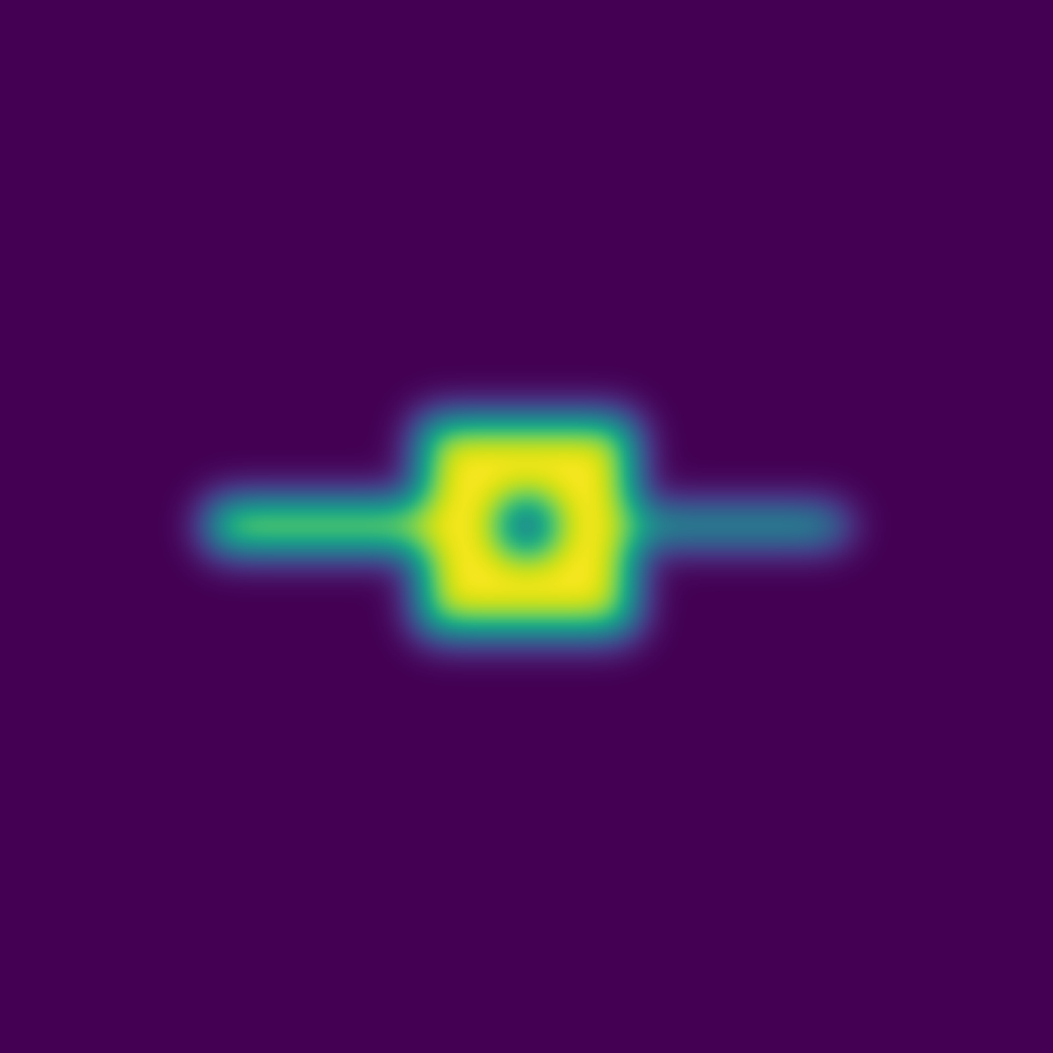};
  
    \nextgroupplot
    \addplot graphics[xmin=0,xmax=1,ymin=0,ymax=1] {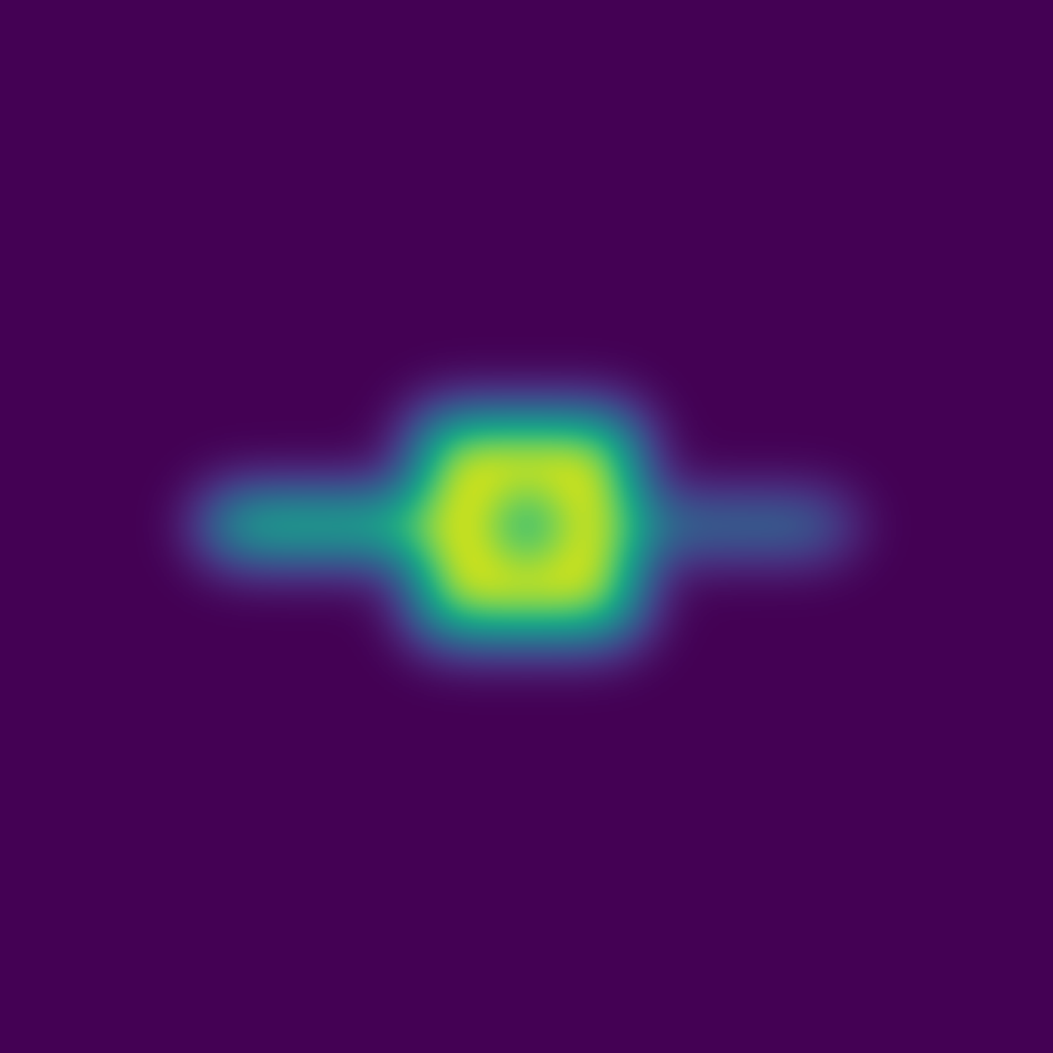};
 
    \nextgroupplot
    \addplot graphics[xmin=0,xmax=1,ymin=0,ymax=1] {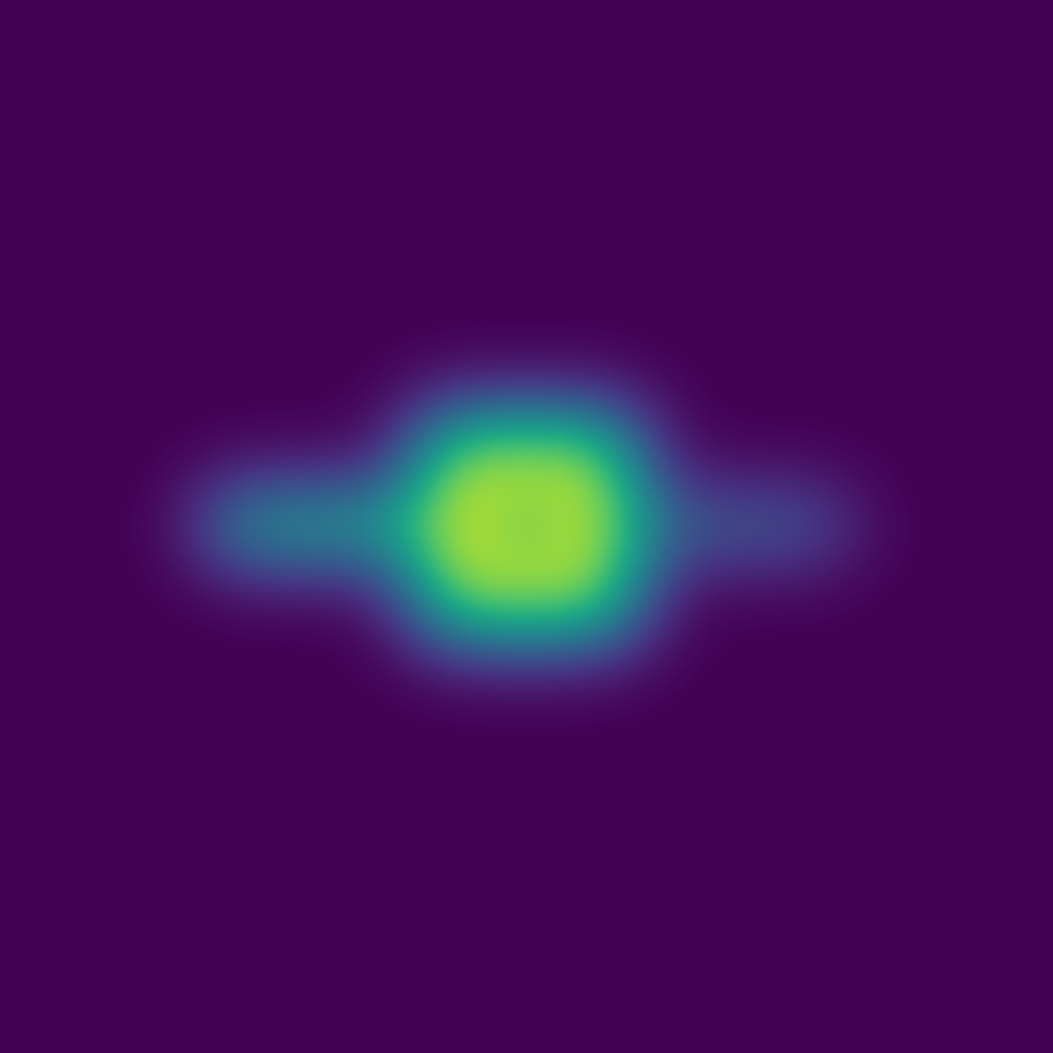};
    
    \nextgroupplot
    \addplot graphics[xmin=0,xmax=1,ymin=0,ymax=1] {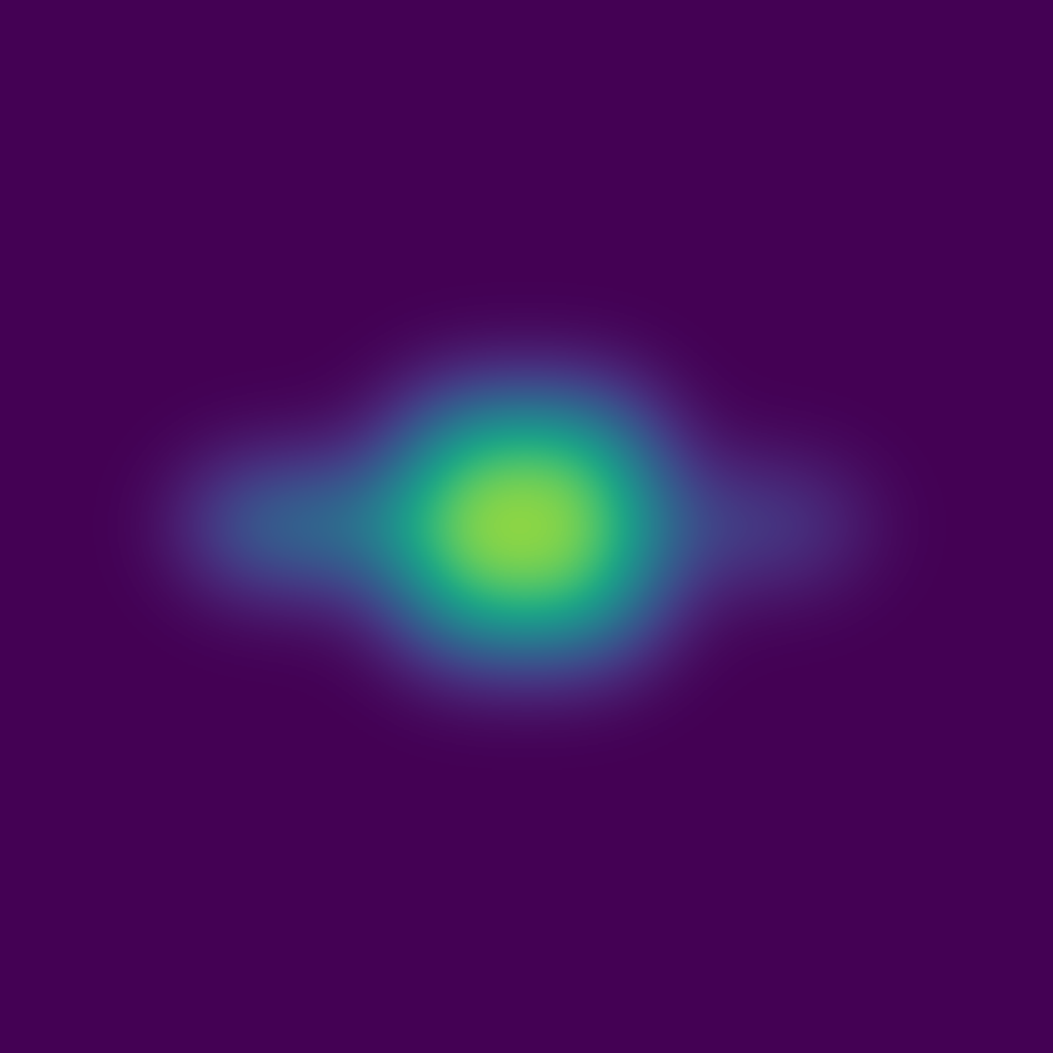};
 
    \nextgroupplot
    \addplot graphics[xmin=0,xmax=1,ymin=0,ymax=1] {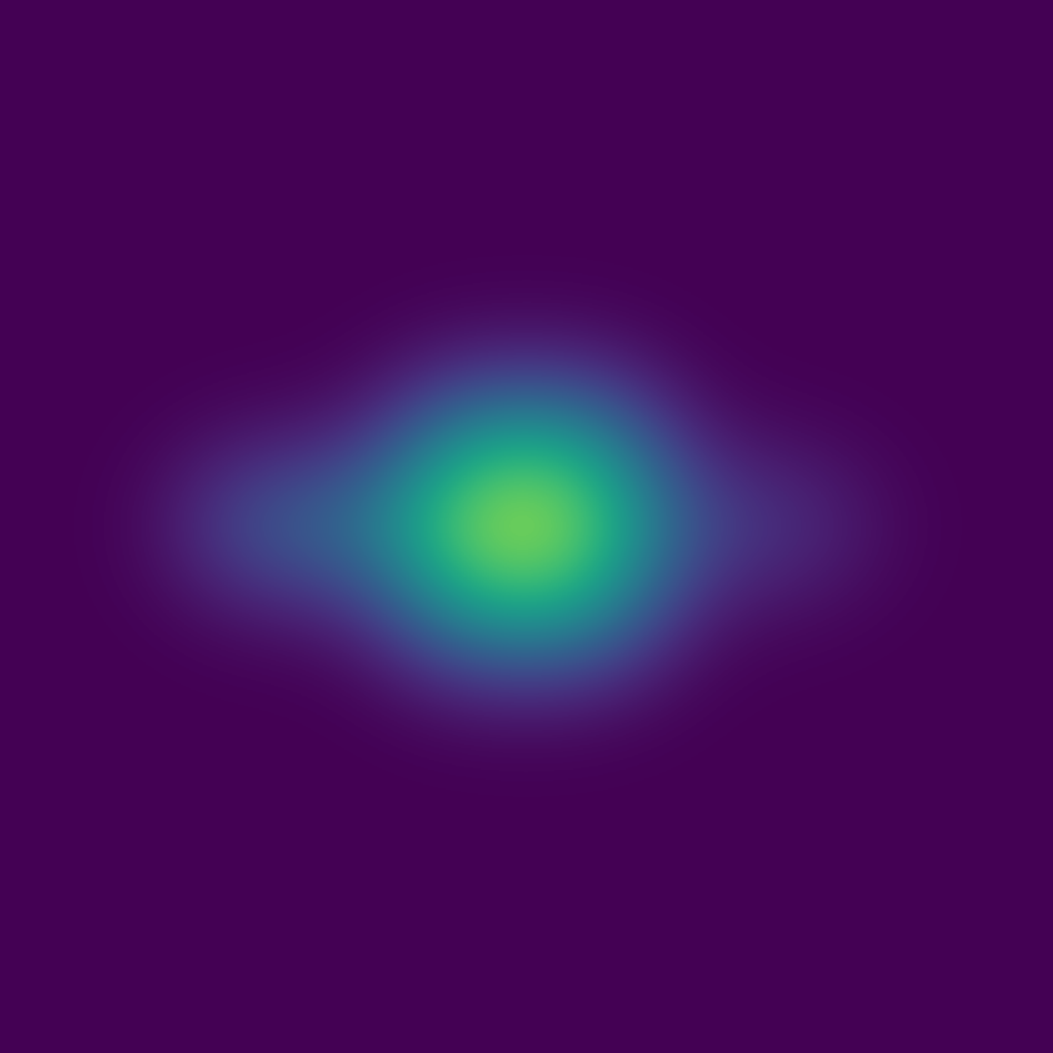};
    
    \nextgroupplot[ylabel={Accuracy}]
    \addplot graphics[xmin=0,xmax=1,ymin=0,ymax=1] {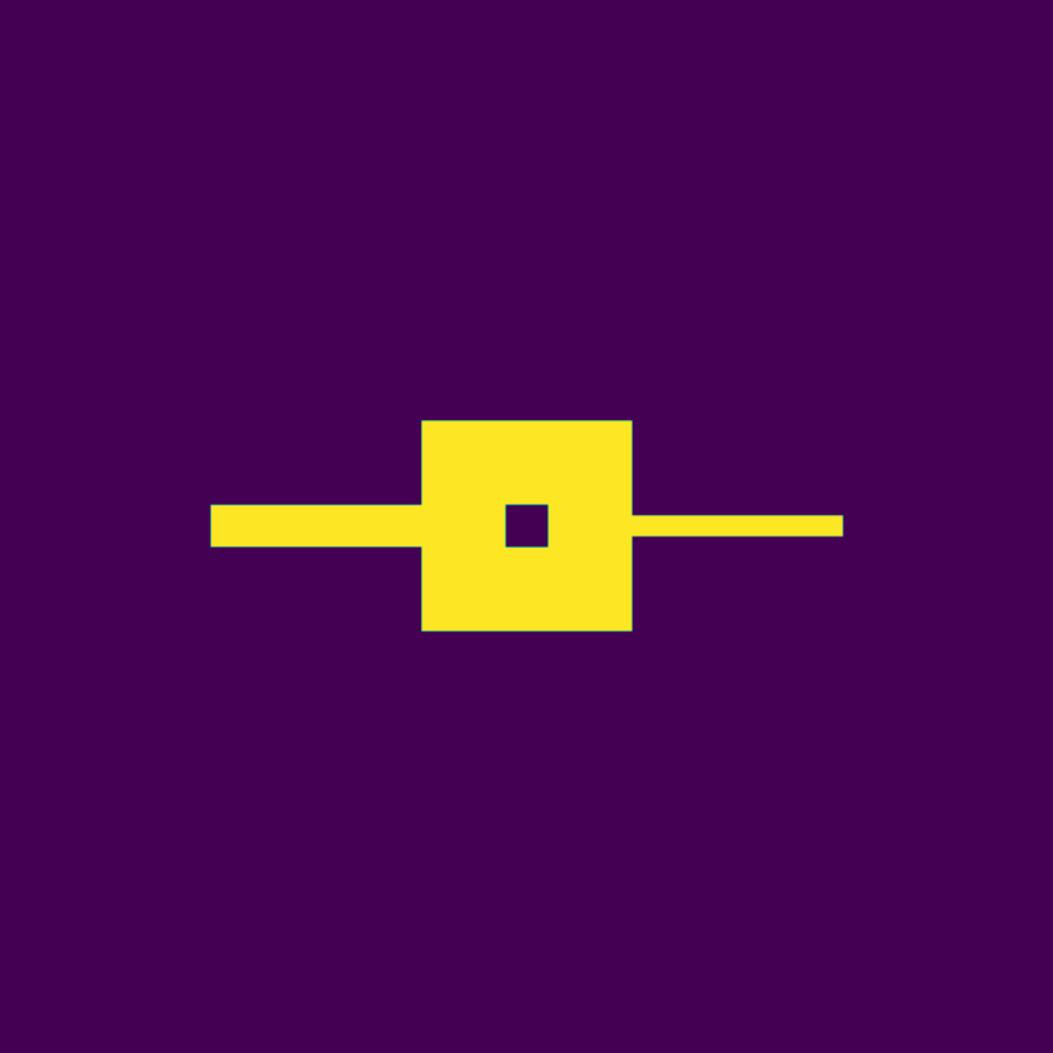};
    
    \nextgroupplot
    \addplot graphics[xmin=0,xmax=1,ymin=0,ymax=1] {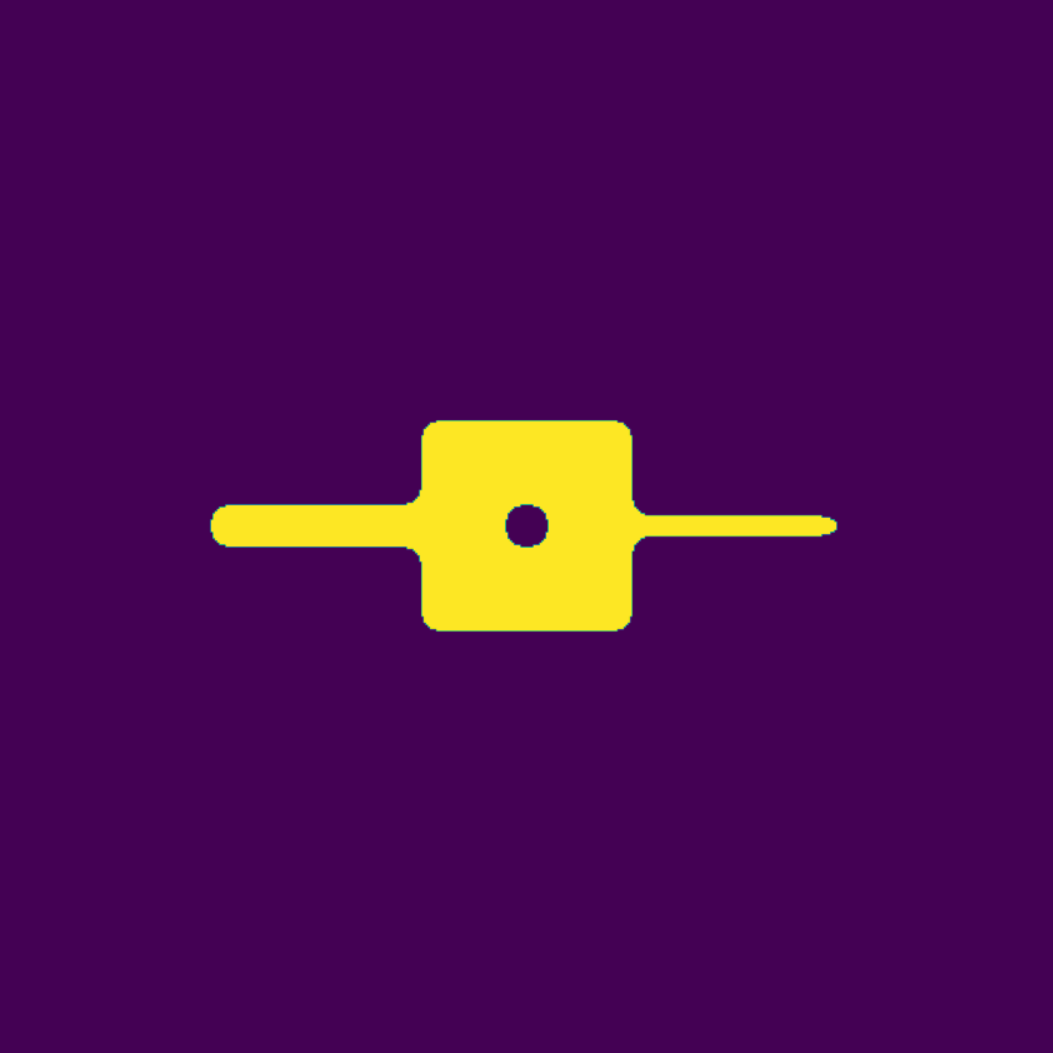};
    
    \nextgroupplot
    \addplot graphics[xmin=0,xmax=1,ymin=0,ymax=1] {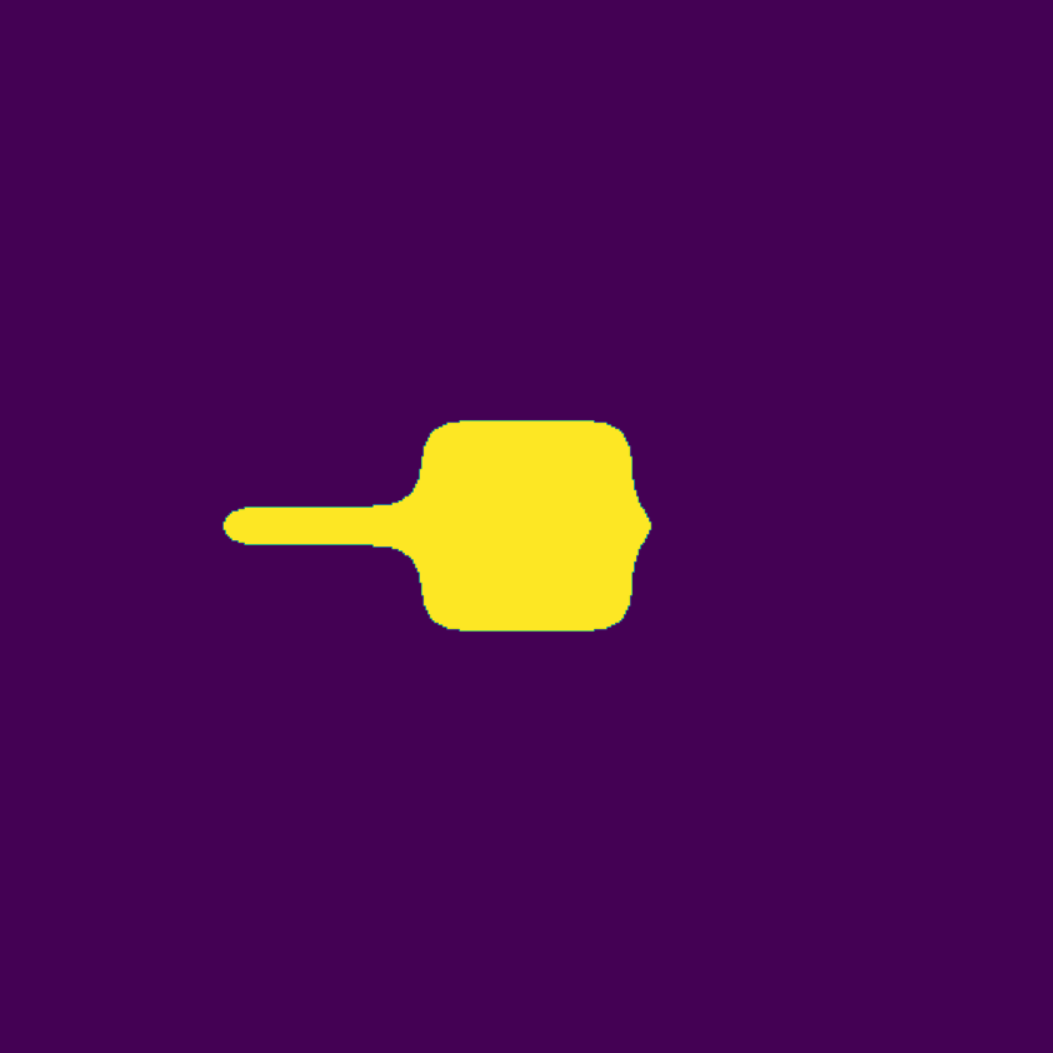};
  
    \nextgroupplot
    \addplot graphics[xmin=0,xmax=1,ymin=0,ymax=1] {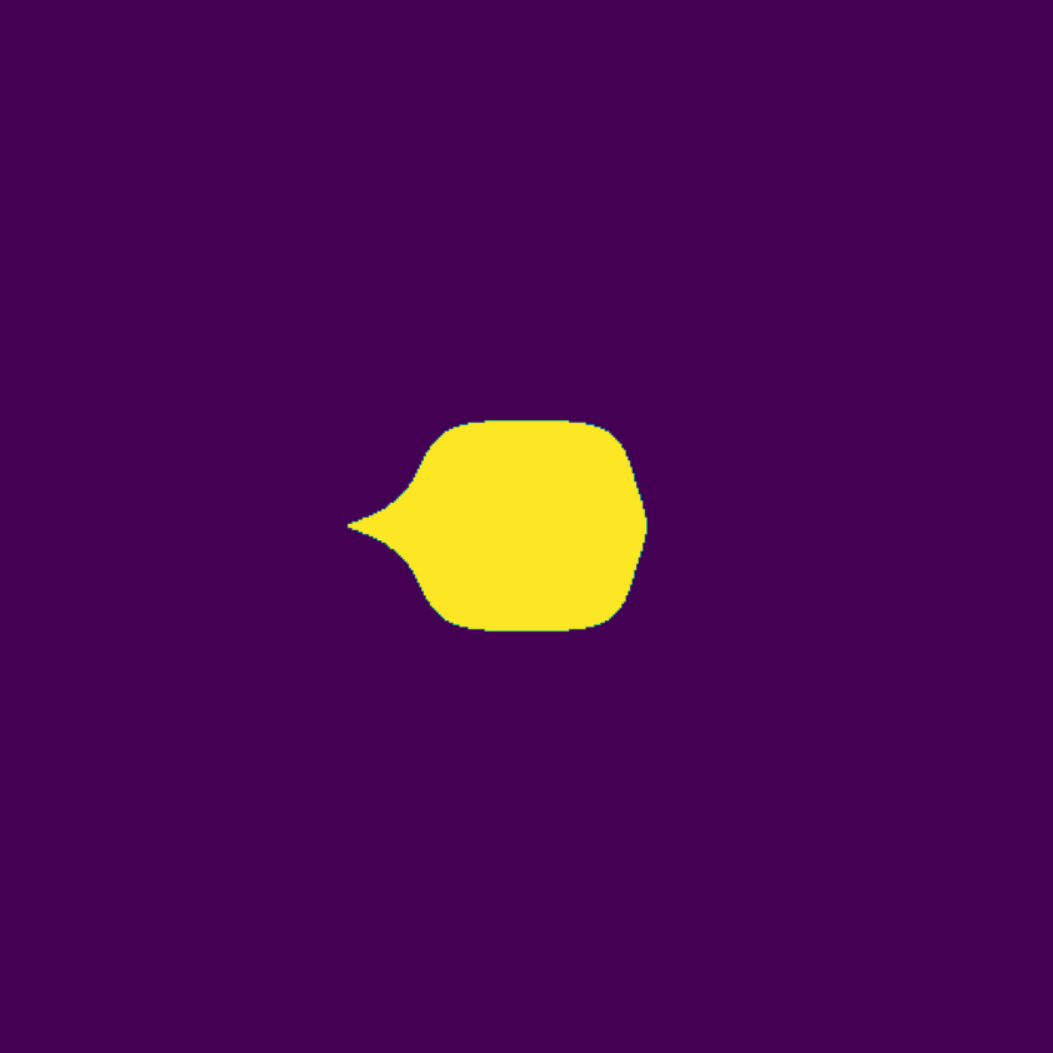};
 
    \nextgroupplot
    \addplot graphics[xmin=0,xmax=1,ymin=0,ymax=1] {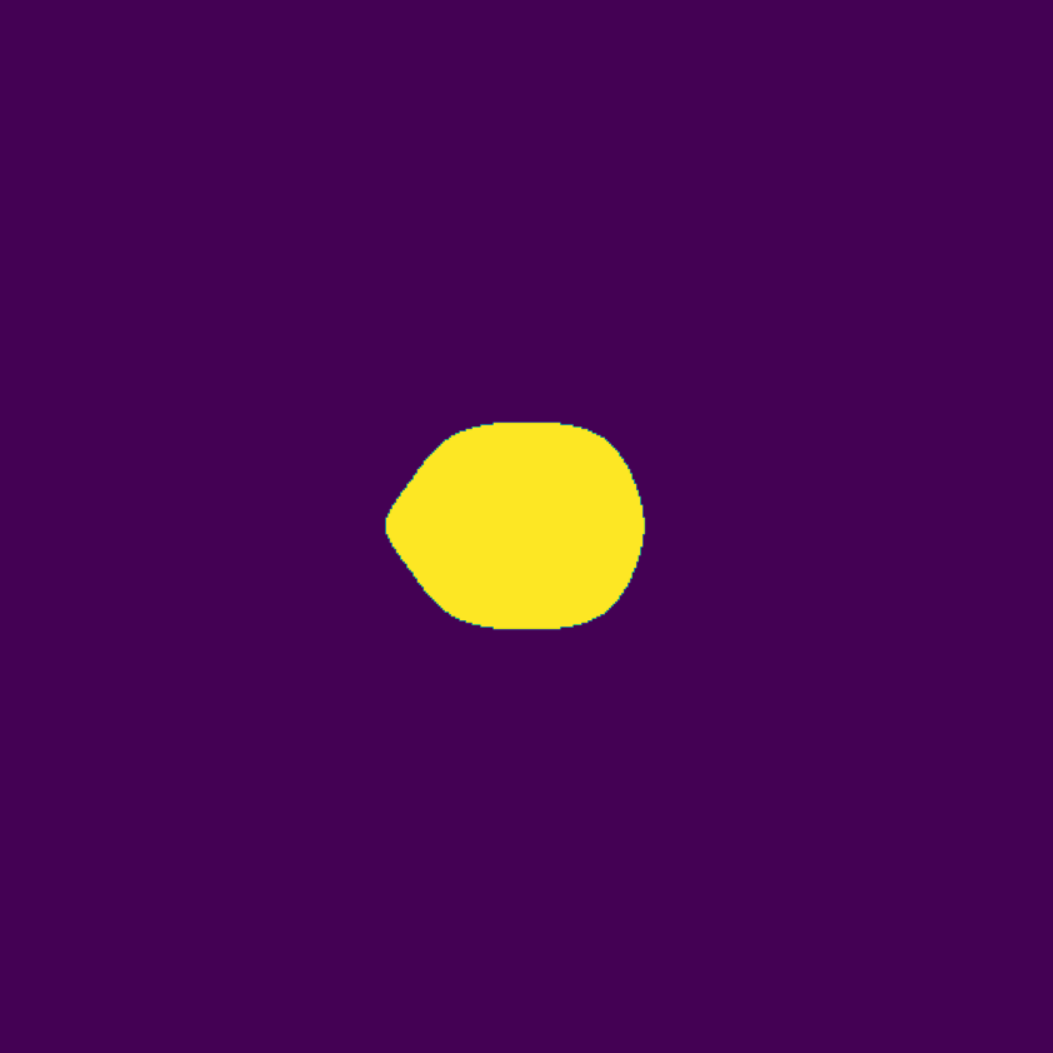};

    \nextgroupplot
    \addplot graphics[xmin=0,xmax=1,ymin=0,ymax=1] {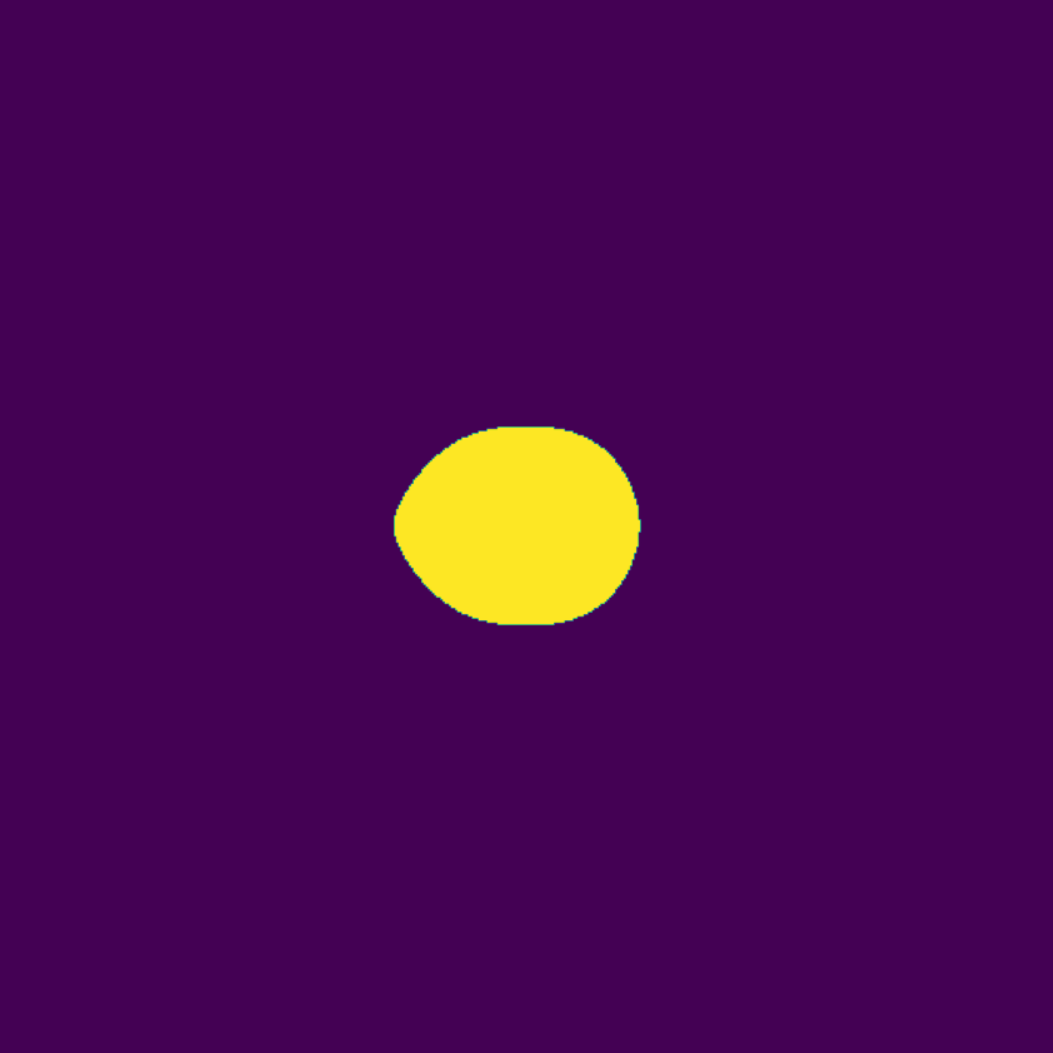};
 
    \nextgroupplot
    \addplot graphics[xmin=0,xmax=1,ymin=0,ymax=1] {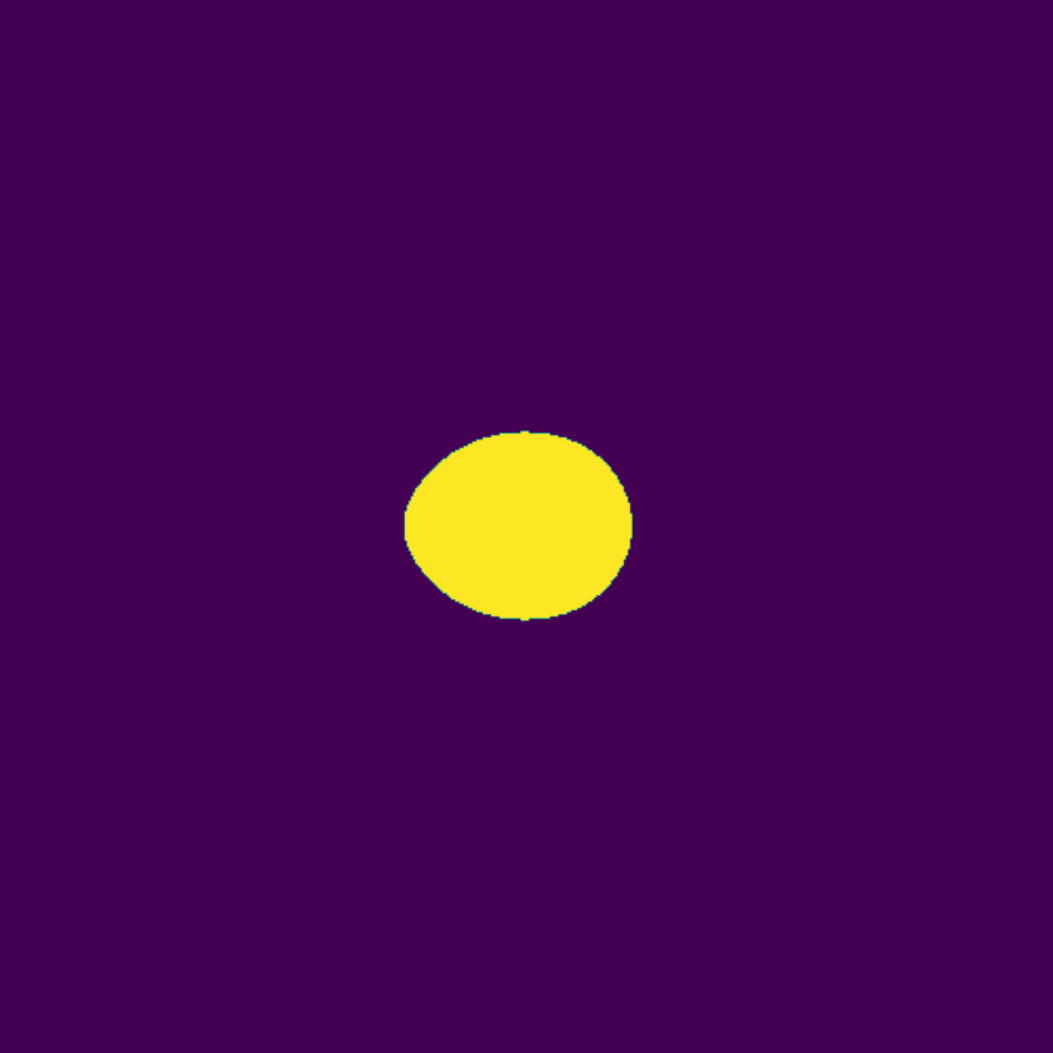};
    
    
    \nextgroupplot[ylabel={Dice}]
    \addplot graphics[xmin=0,xmax=1,ymin=0,ymax=1] {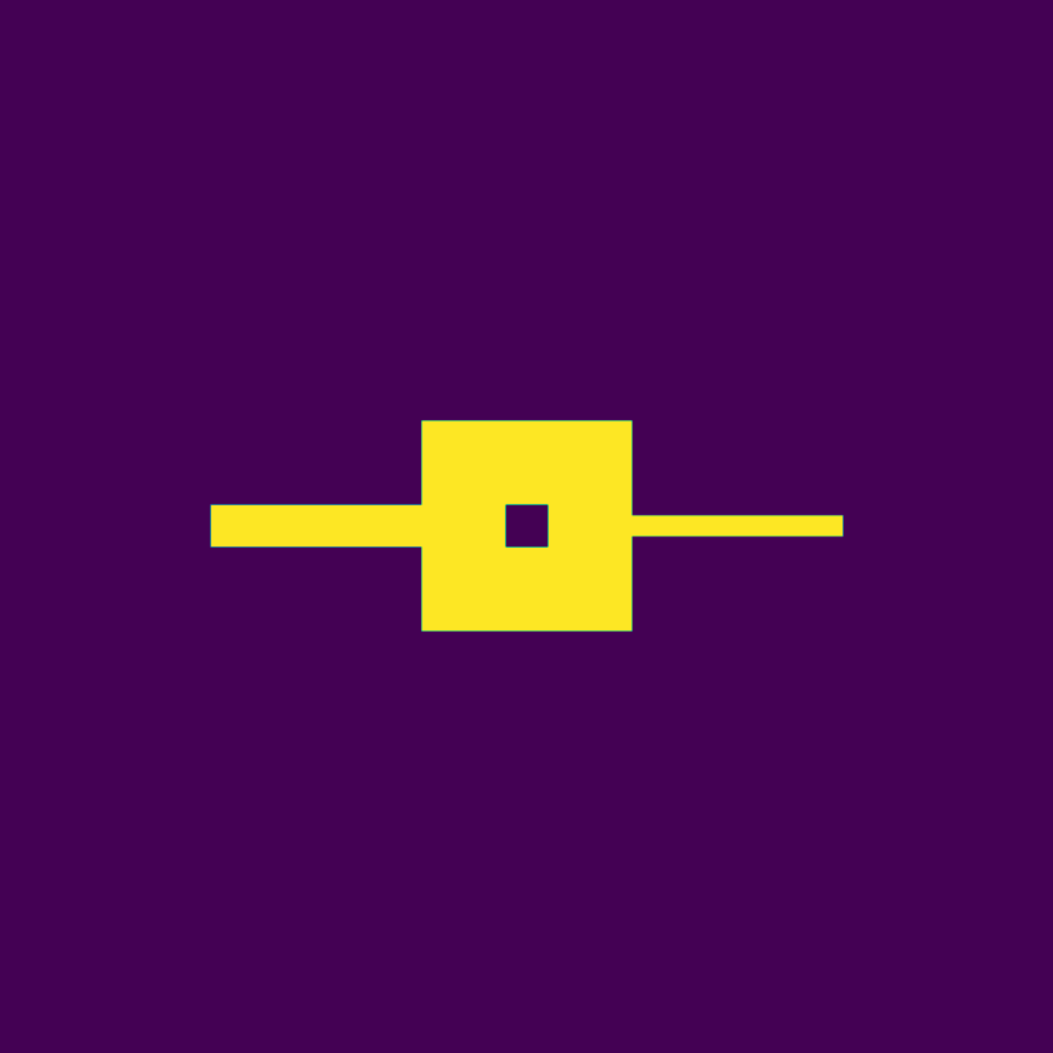};
    
    \nextgroupplot
    \addplot graphics[xmin=0,xmax=1,ymin=0,ymax=1] {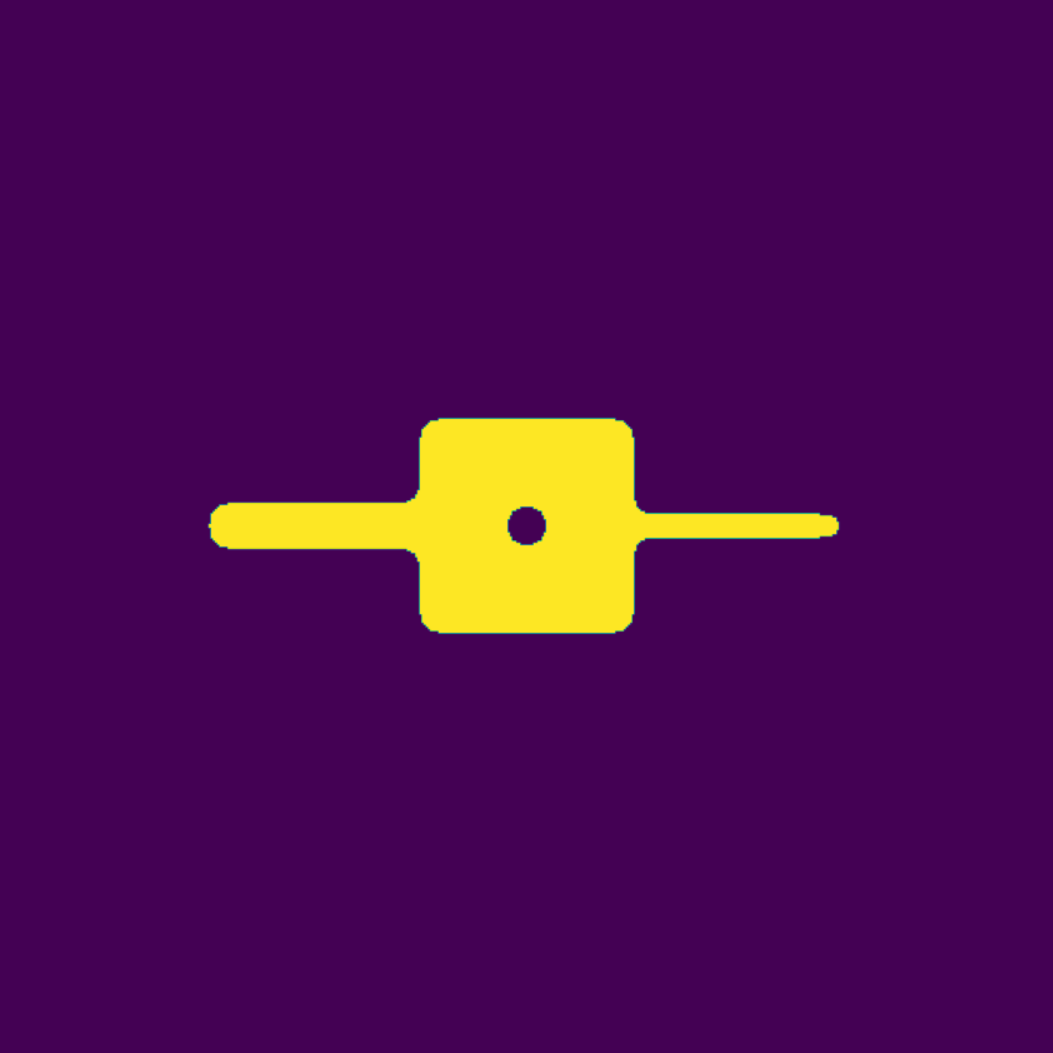};
    
    \nextgroupplot
    \addplot graphics[xmin=0,xmax=1,ymin=0,ymax=1] {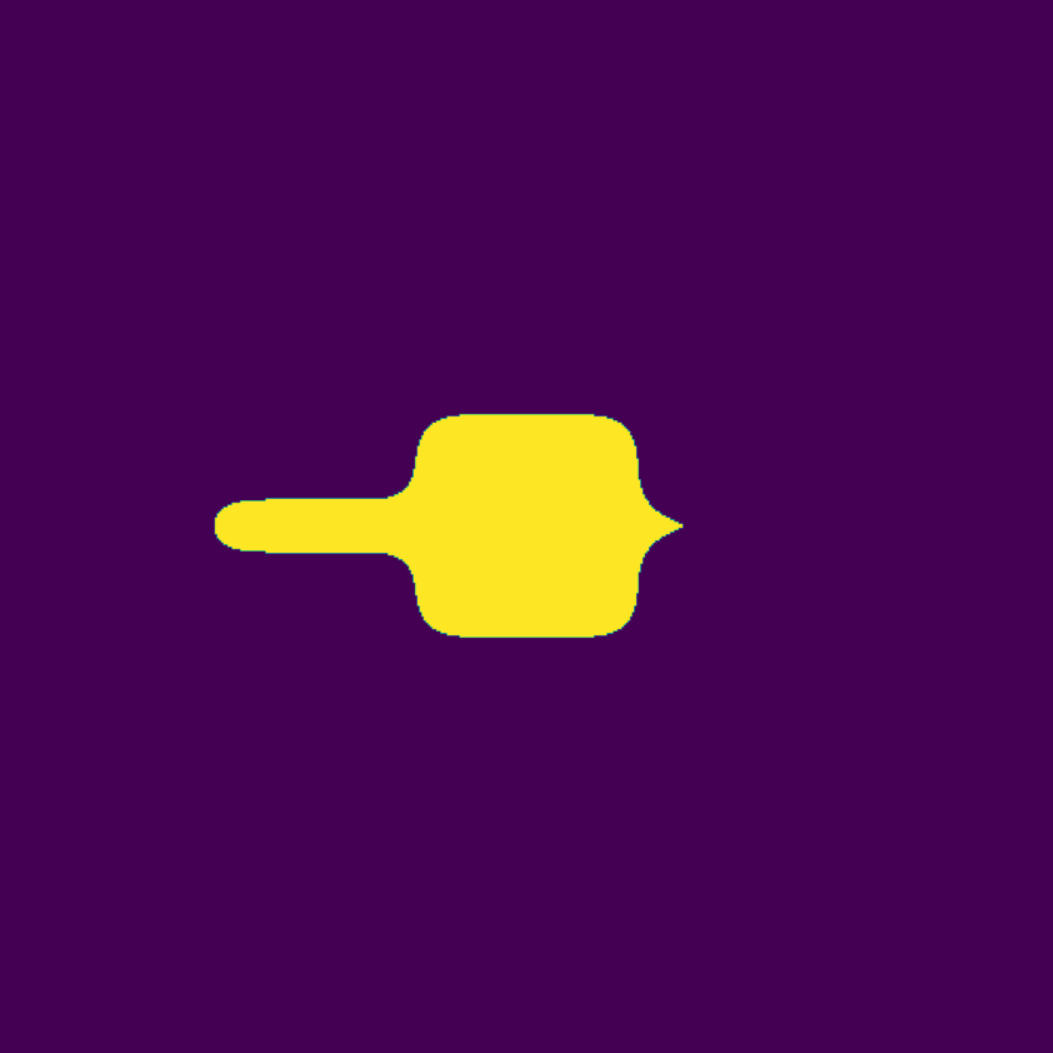};
  
    \nextgroupplot
    \addplot graphics[xmin=0,xmax=1,ymin=0,ymax=1] {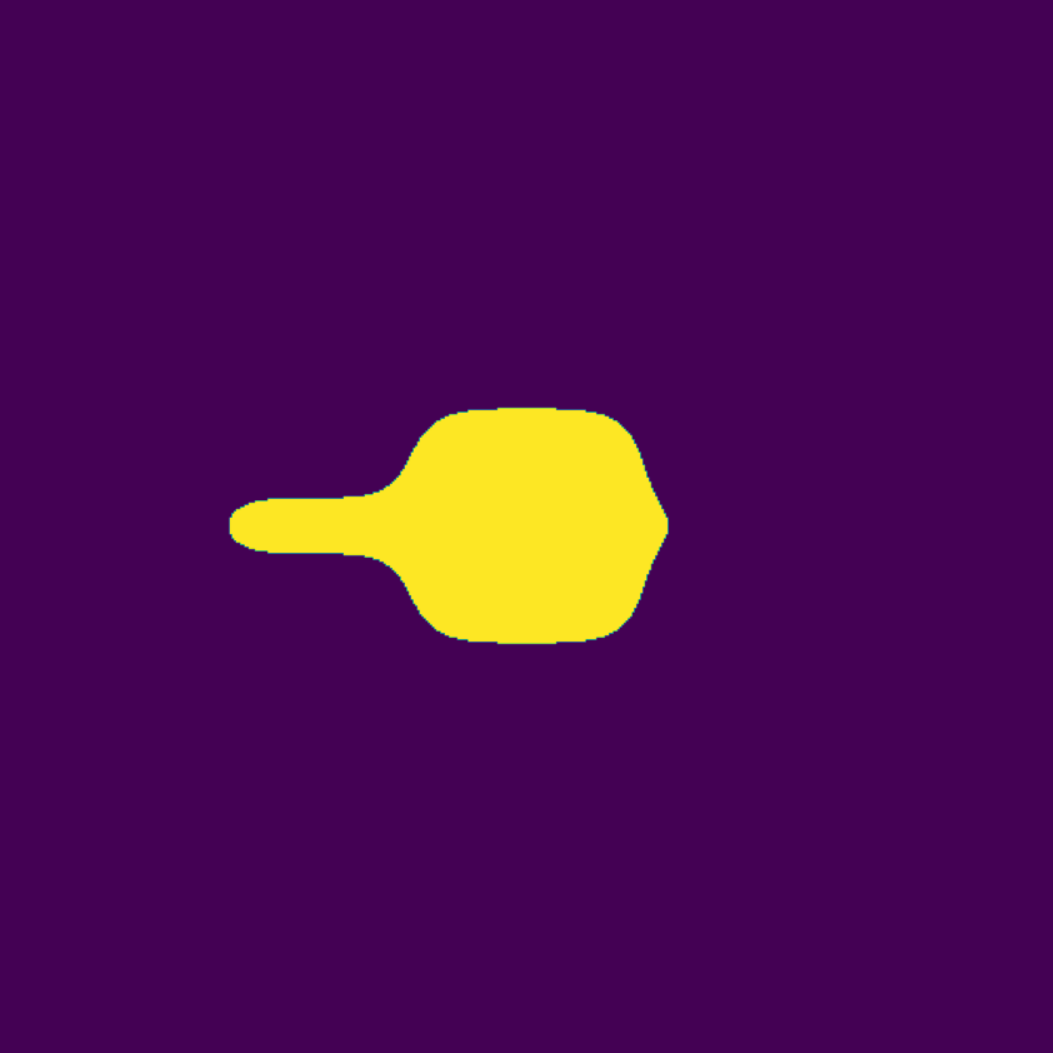};
 
    \nextgroupplot
    \addplot graphics[xmin=0,xmax=1,ymin=0,ymax=1] {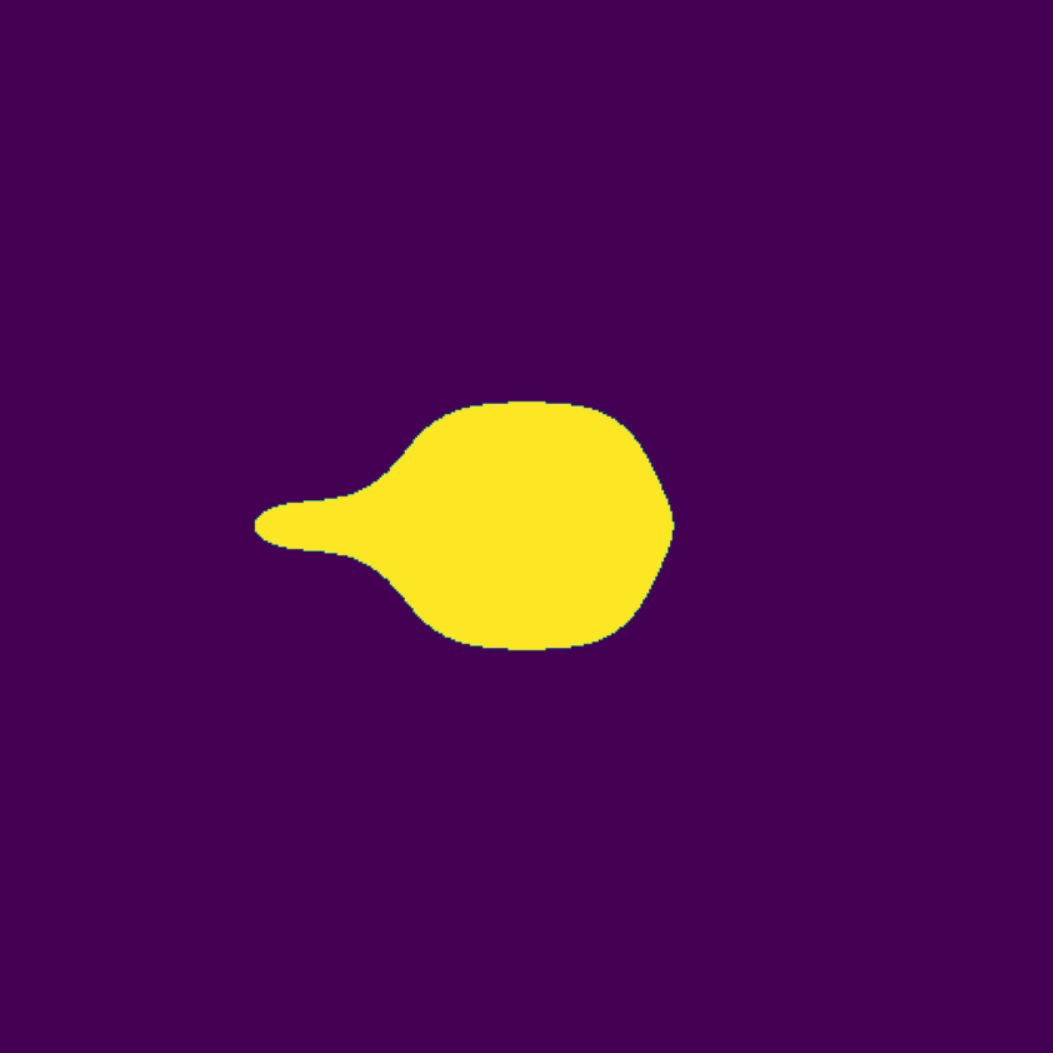};

    \nextgroupplot
    \addplot graphics[xmin=0,xmax=1,ymin=0,ymax=1] {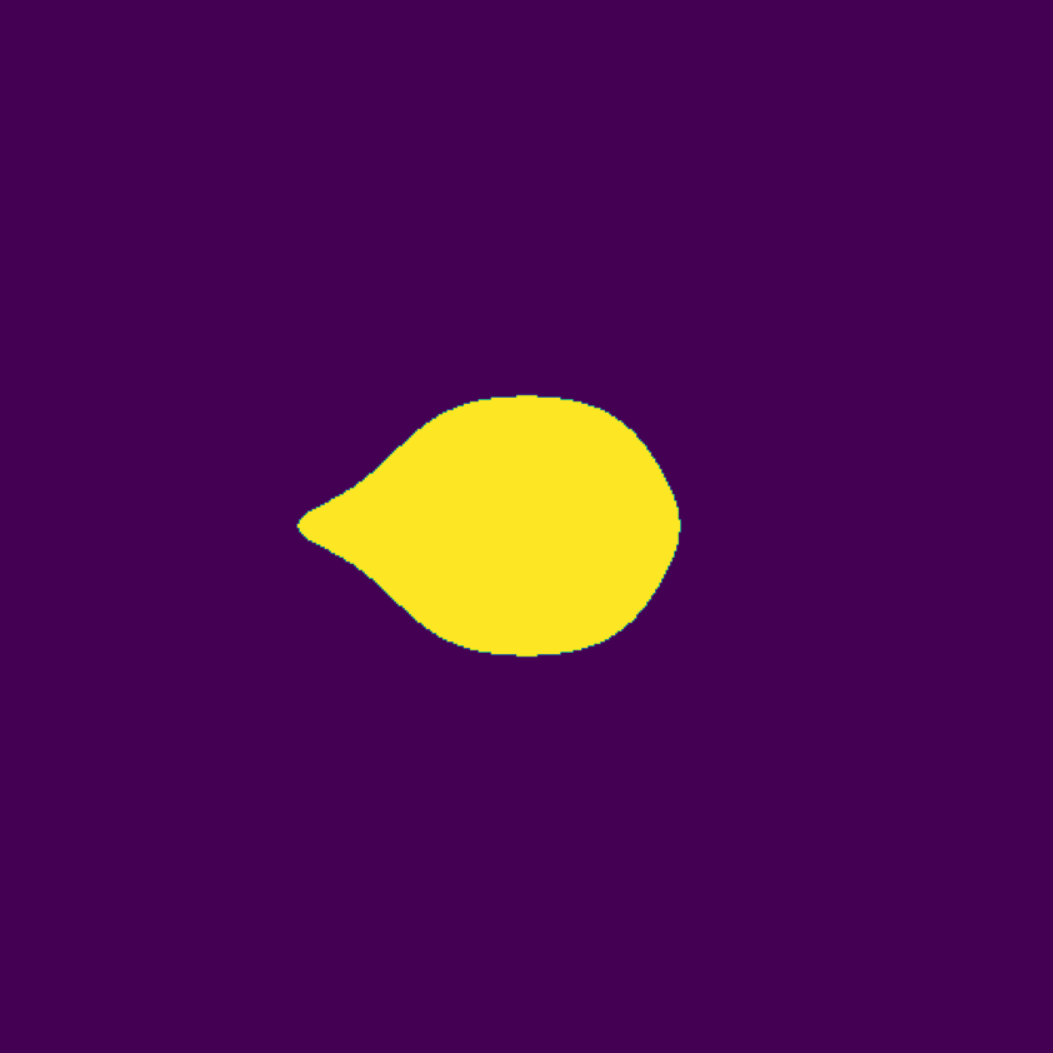};
 
    \nextgroupplot
    \addplot graphics[xmin=0,xmax=1,ymin=0,ymax=1] {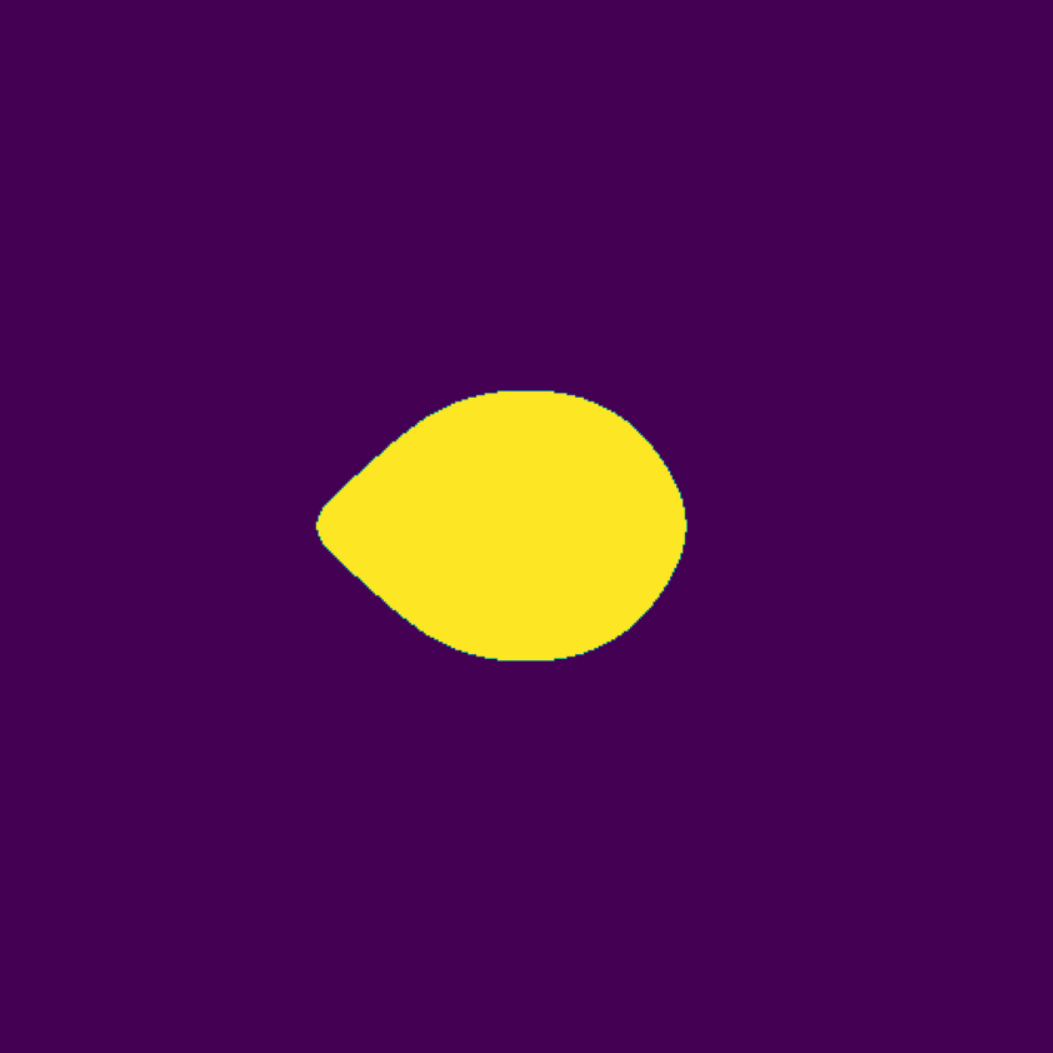};

  \end{groupplot}
\end{tikzpicture}

  \caption{
Illustration of the noise model and its effects on a two dimensional segmentation problem, with the columns showing the effect of different noise strengths.
The first row shows the pixel-wise average obtained from five samples.
The second row shows the marginal function.
The third row shows optimal segmentations associated with Accuracy~\eqref{eq:char_acc}, which in this and most natural cases is unique.
The forth row shows optimal segmentations associated with Dice~\eqref{eq:char_dice}, which in this and most natural cases is unique.
In all of the reported cases $b=0.15/\sqrt{2}$.
}
  \label{fig:ill}
\end{figure*}

\subsection{Optimal segmentations under label noise}
A natural question to ask is in what way the parameters to the noise model affects the optimal solutions obtained for  a loss or metric.
Since the losses and metrics investigated in this work only depend on the marginal function, this question can be answered by firstly addressing in what way the segmentation noise affects the marginal function and secondly addressing in what way the marginal function affects the optimal solutions for a loss or metric.

In Figure~\ref{fig:ill}, these two steps are illustrated for a two dimensional segmentation problem with varying noise intensity. 
In the first row, pixel-wise averages of 5 samples are illustrated.
In the second row, the marginal functions are illustrated.
In the third row, theoretical optimal solutions to Accuracy which in these and most cases are uniquely given are shown.
Note that by~\eqref{eq:ce_cal}, this is also given by the minimizer to cross-entropy thresholded by $1/2$.
In the forth row, theoretical optimal solutions to Dice which in these and most cases are uniquely given are shown.
By~\eqref{eq:sd_cal} and \eqref{eq:cal_thresh}, this is also given by the minimizer to soft-Dice thresholded by $1/2$ and by the minimizer to cross-entropy thresholded by $t_m$~\eqref{eq:threshold}.

A first observation that can be made is that noise affects the optimal segmentations with respect to the two metrics such that the shape is changed as compared to the noise-free structure.
This is an important observation that contradicts the picture that unbiased noise (unbiased in the sense of our model), yield unbiased optimal segmentations.
In particular, both metrics yield segmentations that are changed in such a way that the corners of the shapes are smoothed and details such as thin structures and holes disappear.
Loosely speaking, the higher the noise, the more ball shaped the optimal segmentations become.

A second observation is that the two metrics yield very similar looking segmentations for low noise, but that they yield increasingly different segmentations as the noise increase.
Since cross-entropy with the $1/2$ threshold theoretically gives optimal segmentations with respect to Accuracy and soft-Dice with the $1/2$ threshold theoretically gives optimal segmentations with respect to Dice,
the performance of the two losses as measured by Dice should be very similar in low noise situations but increasingly better for for soft-Dice as the noise is increased.


\begin{figure}
\centering
\begin{tikzpicture}
  \begin{groupplot}[group style={group size=3 by 3, horizontal sep=0.1cm, vertical sep=0.1cm},height=4.25cm,width=4.25cm,xmajorgrids,ymajorgrids,xtick={0.4,0.7,1.0,1.3,1.6},xmin=0.4,xmax=1.6,ytick={0.4,0.7,1.0,1.3,1.6},ymin=0.4,ymax=1.6]

    \nextgroupplot[xtick=\empty,ytick=\empty,yticklabel pos = right, xmin=0, xmax=1,ymin=0,ymax=1]
    \addplot graphics[xmin=0,xmax=1,ymin=0,ymax=1] {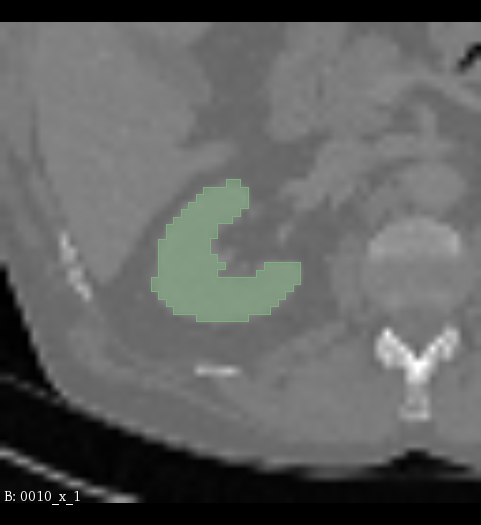};
    
    \nextgroupplot[xtick=\empty,ytick=\empty,yticklabel pos = right, xmin=0, xmax=1,ymin=0,ymax=1]
    \addplot graphics[xmin=0,xmax=1,ymin=0,ymax=1] {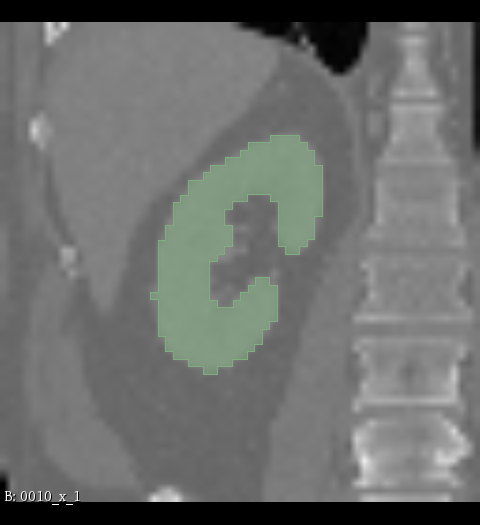};
  
    \nextgroupplot[xtick=\empty,ytick=\empty,yticklabel pos = right, xmin=0, xmax=1,ymin=0,ymax=1]
    \addplot graphics[xmin=0,xmax=1,ymin=0,ymax=1] {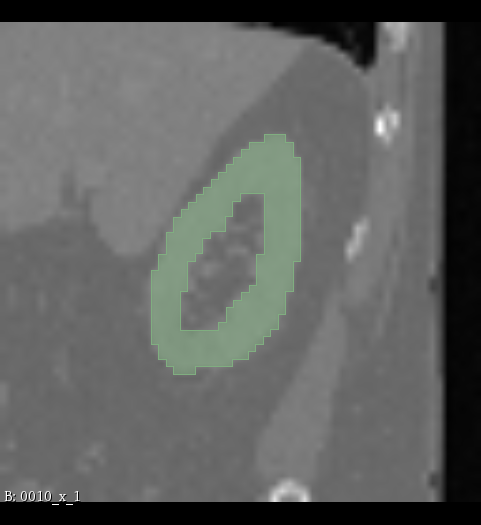};

    
    \nextgroupplot[xtick=\empty,ytick=\empty,yticklabel pos = right, xmin=0, xmax=1,ymin=0,ymax=1]
    \addplot graphics[xmin=0,xmax=1,ymin=0,ymax=1] {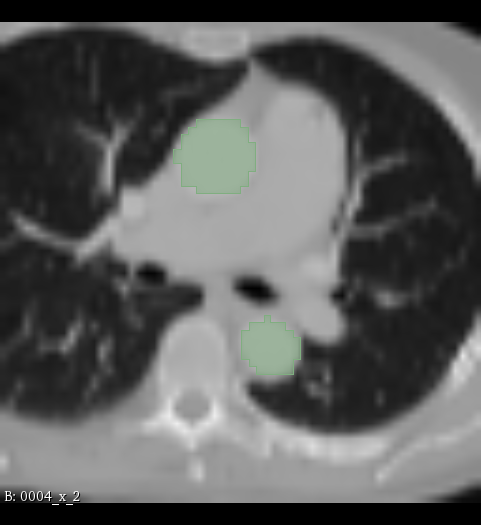};
    
    \nextgroupplot[xtick=\empty,ytick=\empty,yticklabel pos = right, xmin=0, xmax=1,ymin=0,ymax=1]
    \addplot graphics[xmin=0,xmax=1,ymin=0,ymax=1] {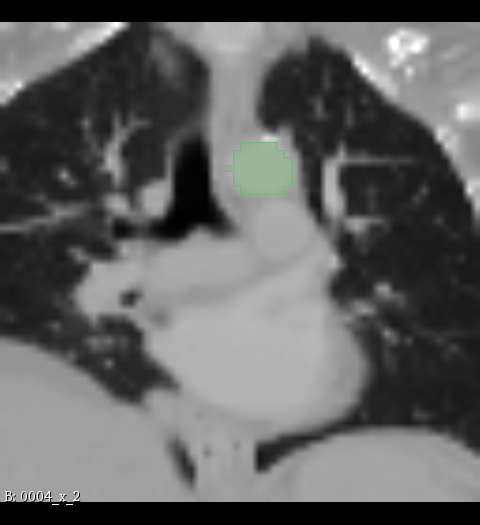};
  
    \nextgroupplot[xtick=\empty,ytick=\empty,yticklabel pos = right, xmin=0, xmax=1,ymin=0,ymax=1]
    \addplot graphics[xmin=0,xmax=1,ymin=0,ymax=1] {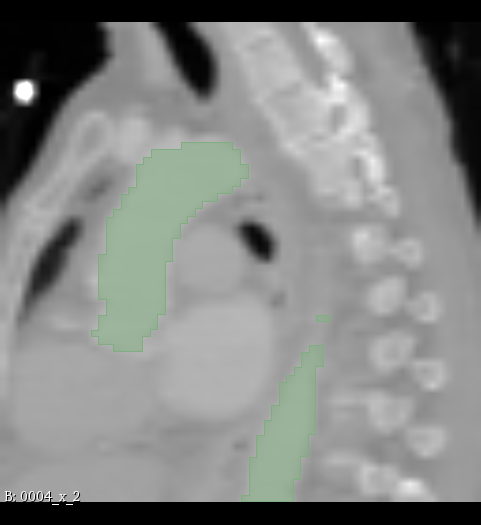};
    
    \nextgroupplot[xtick=\empty,ytick=\empty,yticklabel pos = right, xmin=0, xmax=1,ymin=0,ymax=1]
    \addplot graphics[xmin=0,xmax=1,ymin=0,ymax=1] {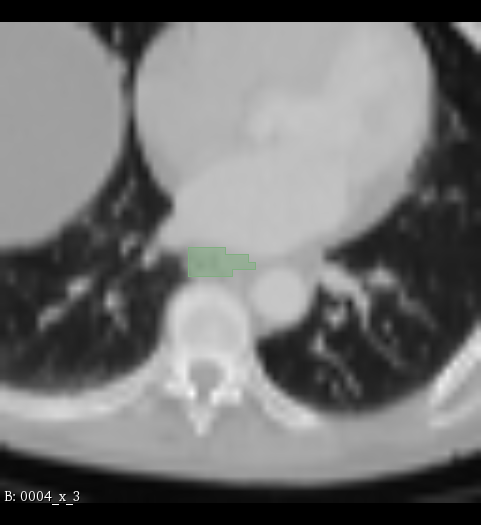};
    
    \nextgroupplot[xtick=\empty,ytick=\empty,yticklabel pos = right, xmin=0, xmax=1,ymin=0,ymax=1]
    \addplot graphics[xmin=0,xmax=1,ymin=0,ymax=1] {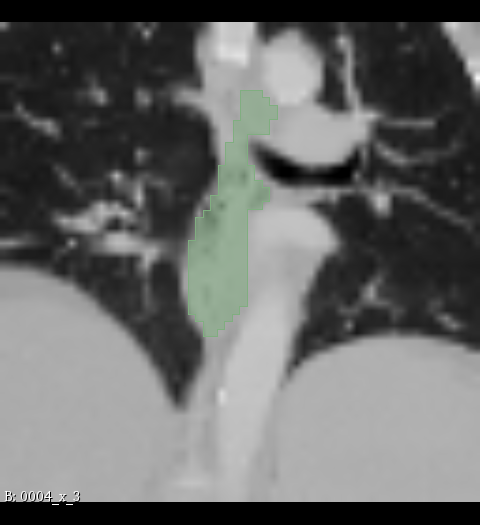};
  
    \nextgroupplot[xtick=\empty,ytick=\empty,yticklabel pos = right, xmin=0, xmax=1,ymin=0,ymax=1]
    \addplot graphics[xmin=0,xmax=1,ymin=0,ymax=1] {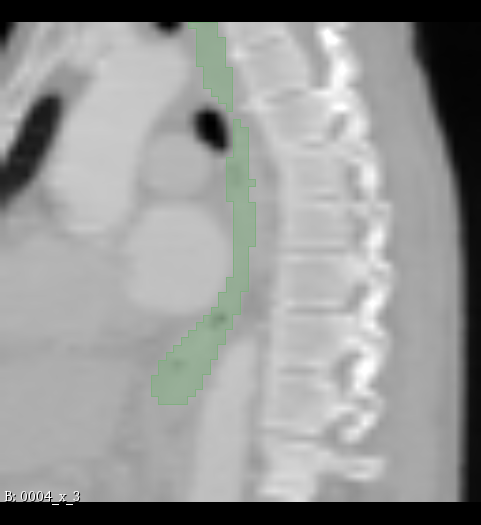};

  \end{groupplot}
\end{tikzpicture}

  \label{seqfig}

  \caption{
  Illustration of the data from the TotalSegmentor data set~\cite{wasserthal2022totalsegmentator} used in the experiments.
  The first row shows the axial, coronal and sagittal views passing the center of mass of the kidney in one of the data points.
  The second row shows the axial, coronal and sagittal views passing the center of mass of the aorta in one of the data points. 
  The third row shows the axial, coronal and sagittal views passing the center of mass of the esophagus in one of the data points.
  }
\end{figure}

\section{Experiments}
In this section the three discussed methods are experimentally tested using the derived noise model.
For short, the notation $\mathrm{CE}^{(0)}$ will be used to denote a model trained with cross-entropy and thresholded with $1/2$, $\mathrm{SD}^{(0)}$ a model trained with cross-entropy and thresholded with $1/2$ and $\mathrm{CE}^{(*)}$ a model trained with cross-entropy and thresholded according to~\eqref{eq:threshold}.

\subsection{Data}
The experiments are conducted with respect to the TotalSegmentor data set~\cite{wasserthal2022totalsegmentator}.
This data set contains 1204 CT images with 104 anatomical structures (27 organs, 59 bones, 10 muscles, 8
vessels).
To illustrate the effect noise may have on organs with different shape, three different organs are chosen.
This includes the right kidney which in general is pretty spherical, the aorta which is tubular and relatively thick and the esophagus which is tubular and relatively thin.
For each of the organs, 400 cases is selected and split into 5 folds of 80 cases.
Finally, the images are sub-sample to half resolution and patches of $64^3$ voxels centered in each of the structures are extracted.

\subsection{Model}
A standard 3D U-net composed of blocks of two convolutions with instance normalization and relu-activations is implemented.
The number of features used increases for each down-sampling according to $16,32,64,128,256$, where $16$ corresponds to the input resolution and $256$ the bottleneck.
Convolutions with kernel $2^3$ stride $2$ are used for down-sampling and
and transposed convolutions with kernel $2^3$ stride $2$ are used for up-sampling.

\subsection{Training}
The models are trained with either cross-entropy or soft-Dice, in both cases using $l_2$ regularization with a weight of $10^{-5}$.
Adam with learning rate of $10^{-4}$ is used for optimization together with a scheduler that divides the learning rate by 5 after 30 epochs of non-improvement.
The number of epochs computed is fixed to 100 and the order of the training data is randomly shuffled.
Batch sizes of $1$ is used, and with probability $0.5$ the data is augmented.
The augmentation is composed of random shifts, random rotations, applying Gaussian filters with random smoothing levels of smoothing.
Performance is measured with respect to the model that achieves the highest metric score on the training data.

\subsection{Results}
\begin{table}
    \centering

    \begin{tabular}{l@{\hspace{0.6cm}}c@{\hspace{0.6cm}}c@{\hspace{0.6cm}}c@{\hspace{0.6cm}}c}
\toprule
Organ & $a$ & $\mathrm{CE}^{(0)}$ &  $\mathrm{SD}^{(0)}$  & $\mathrm{CE}^{(*)}$ \\
\midrule
Kidney
& 0.0000 & 0.9611 & 0.9634 & 0.9615 \\
& 0.0100 & 0.8762 & 0.8794 & 0.8774 \\
& 0.0200 & 0.7883 & 0.7914 & 0.7909 \\
& 0.0300 & 0.6947 & 0.7080 & 0.7055 \\
\midrule
Aorta
& 0.0000 & 0.9525 & 0.9515 & 0.9524 \\
& 0.0100 & 0.8639 & 0.8654 & 0.8653 \\
& 0.0200 & 0.7557 & 0.7569 & 0.7600 \\
& 0.0300 & 0.6215 & 0.6513 & 0.6560 \\
\midrule
Esophagus
& 0.0000 & 0.8552 & 0.8603 & 0.8602 \\
& 0.0100 & 0.6671 & 0.6722 & 0.6814 \\
& 0.0200 & 0.4168 & 0.4527 & 0.4829 \\
& 0.0300 & 0.1441 & 0.3105 & 0.3489 \\
\bottomrule
\end{tabular}
\caption{Table over the results from the experiments for each organ and noise level $a$.
The entries are average Dice scores obtained from the average over the five folds.
$\mathrm{CE}^{(0)}$ indicate that cross-entropy with $1/2$ threshold has been used, $\mathrm{SD}^{(0)}$ indicate that soft-Dice with a $1/2$-threshold has been used and $\mathrm{CE}^{(*)}$ indicate that cross-entropy with the threshold described in~\eqref{eq:threshold} has been used.}
\label{table:results}
\end{table}

%

\begin{figure*}
\begin{tikzpicture}
  \begin{groupplot}[group style={group size=3 by 3, horizontal sep=1.2cm},height=6.1cm,width=6.1cm,xticklabel style={
            /pgf/number format/fixed,
            /pgf/number format/precision=2,
            /pgf/number format/fixed zerofill
        }, scaled x ticks=false]
    
    \nextgroupplot[title=Kidney,xlabel=$a$,ylabel=Dice]
    \addplot[color=red,mark=x] coordinates {
        (0.00,0.9611)
        (0.01,0.8762)
        (0.02,0.7883)
        (0.03,0.6947)
    };
    \addlegendentry{$\mathrm{CE}^{(0)}$}
    
    \addplot[color=blue,mark=x] coordinates {
        (0.00,0.9634)
        (0.01,0.8794)
        (0.02,0.7914)
        (0.03,0.7080)
    };
    \addlegendentry{$\mathrm{SD}^{(0)}$}
    
    \addplot[color=green,mark=x] coordinates {
        (0.00,0.9615)
        (0.01,0.8774)
        (0.02,0.7909)
        (0.03,0.7055)
    };
    \addlegendentry{$\mathrm{CE}^{(*)}$}

    \nextgroupplot[title=Aorta,xlabel=$a$,ylabel=Dice]
    \addplot[color=red,mark=x] coordinates {
        (0.00,0.9525)
        (0.01,0.8639)
        (0.02,0.7557)
        (0.03,0.6215)
    };
    \addlegendentry{$\mathrm{CE}^{(0)}$}
    
    \addplot[color=blue,mark=x] coordinates {
        (0.00,0.9515)
        (0.01,0.8654)
        (0.02,0.7569)
        (0.03,0.6513)
    };
    \addlegendentry{$\mathrm{SD}^{(0)}$}
    
    \addplot[color=green,mark=x] coordinates {
        (0.00,0.9524)
        (0.01,0.8653)
        (0.02,0.7600)
        (0.03,0.6560)
    };
    \addlegendentry{$\mathrm{CE}^{(*)}$}

    \nextgroupplot[title=Esophagus,xlabel=$a$,ylabel=Dice]
    \addplot[color=red,mark=x] coordinates {
        (0.00,0.8552)
        (0.01,0.6671)
        (0.02,0.4168)
        (0.03,0.1441)
    };
    \addlegendentry{$\mathrm{CE}^{(0)}$}
    
    \addplot[color=blue,mark=x] coordinates {
        (0.00,0.8603)
        (0.01,0.6722)
        (0.02,0.4527)
        (0.03,0.3105)
    };
    \addlegendentry{$\mathrm{SD}^{(0)}$}
    
    \addplot[color=green,mark=x] coordinates {
        (0.00,0.8602)
        (0.01,0.6814)
        (0.02,0.4829)
        (0.03,0.3489)
    };
    \addlegendentry{$\mathrm{CE}^{(*)}$}
    
  \end{groupplot}
\end{tikzpicture}

\caption{Illustration of the results of the experiments presented in Table~\ref{table:results}.
To the left are the results from the kidney segmentation problem, in the middle are the results from the aorta segmentation problem and to the right are the results from the esophagus segmentation problem.
In each of the figures, the average Dice value obtained from five fold experiments is plotted as a function of the noise level $a$.
The legend $\mathrm{CE}^{(0)}$ indicate that cross-entropy with the $1/2$-threshold has been used,
 $\mathrm{SD}^{(0)}$ indicate that soft-Dice with the $1/2$-threshold has been used and $\mathrm{CE}^{(*)}$ indicate that cross-entropy with the threshold described in ~\eqref{eq:threshold} has been used.
 In all of the reported cases $b=0.15/\sqrt{2}$.
}

\label{fig:results}
\end{figure*}
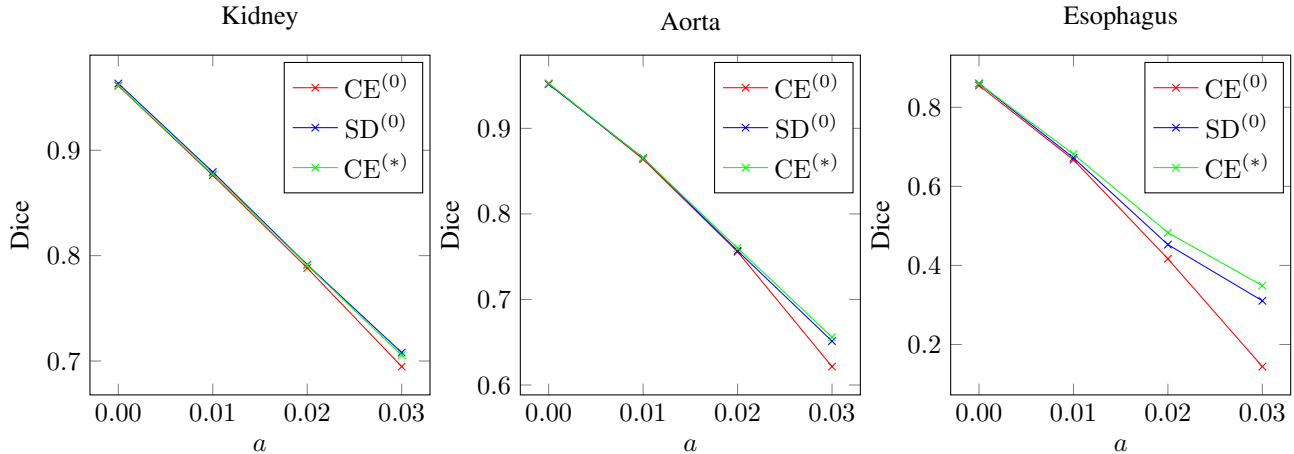
The results from the experiments are shown in Table~\ref{table:results} and Figure~\ref{fig:results}.
This includes the average Dice score obtained for the five different folds and four different noise levels, for each of the different methods of obtaining segmentations: $\mathrm{CE}^{(0)}$, $\mathrm{SD}^{(0)}$ and $\mathrm{CE}^{(*)}$.

The first thing to notice is that either all of the the methods perform very similarly, or $\mathrm{SD}^{(0)}$ threshold performs better than $\mathrm{CE}^{(0)}$.
Also, whenever $\mathrm{SD}^{(0)}$ performs better than $\mathrm{CE}^{(*)}$, cross-entropy with the Dice optimal threshold performs similarly or better than $\mathrm{SD}^{(0)}$.
Simply put, using the alternative threshold~\eqref{eq:threshold} with cross-entropy yields the performance of soft-Dice.
The second thing to notice is that the difference amongst the methods increases when noise increases.
That is, for small noise, the performance of the methods are almost equivalent, but as the noise increases, the associated scores start to diverge.
This type of result is expected from the theory and depicted in Figure~\ref{fig:ill}.

\section{Discussion}

The experiments provided in this paper systematically validated the theoretical observations.
However, the score as measured by the average Dice value, is very low before the effects start to show.
In previous work comparing soft-Dice and cross-entropy the effect seems to follow this pattern in the sense that the difference is larger when the score is lower~\cite{bertels2019optimizing}.
However, there are situations when the Dice score is high but with a significant difference.
This could be for many reasons, but one possibility is that there exist some other noise that has this property.
In general however, the results indicate that there is evidence for using cross-entropy rather than soft-Dice, provided the right threshold is used.
This is also preferred for several other reasons, such as the better stability properties of cross-entropy and that it provides marginal estimates that can be useful.

\section{Conclusion}
In this work, the optimal solutions to soft-Dice and cross-entropy has been compared and it is illustrated that optimal solutions to soft-Dice can be obtained by thresholding cross-entropy using computable threshold in general different from $1/2$.
A realistic label noise model based on Gaussian fields is proposed and explicit formulations for marginal functions are derived.
Also, an efficient method for sampling noisy segmentations is derived.
Using the noise model, it is illustrated how $1/2$-thresholded optimal solutions to cross-entropy and soft-Dice are affected by various noise levels, and that the resulting segmentations diverge as the noise is increased.
It is shown that optimal solutions to soft-Dice can be found by thresholding the optimal solutions to cross-entropy with a computable a priori unknown threshold than $1/2$.
Finally, the theoretical observations is verified on real data from the TotalSegmentor data set~\cite{wasserthal2022totalsegmentator}.

\paragraph{Limitations:}
There are two main limitations in this work.
Firstly, the theoretical results only hold exactly when the volume of the noisy label is constant.
Secondly, it is assumed that the noise can be modeled with the proposed Gaussian field model.
In practice, noise will not follow such a model exactly since it will often be the case that the amount of noise will differ for different regions of the target structure.

\paragraph{Acknowledgement:}
The work was partially funded by RaySearch Laboratories AB.

\clearpage
{\small
\bibliographystyle{ieee_fullname}
\bibliography{bibliography}
}

\end{document}


\onecolumn

\title{--- Supplementary Document --- \\ Marginal Thresholding in Noisy Image Segmentation}

\author{
  Marcus Nordstr\"om \thanks{Author is also affiliated with RaySearch Laboratories.}\\
  Department of Mathematics\\
  KTH Royal Institute of Technology\\
  Stockholm, Sweden \\
  {\tt\small marcno@kth.se} \\
  \and
  Henrik Hult \\
  Department of Mathematics\\
  KTH Royal Institute of Technology\\
  Stockholm, Sweden \\
  {\tt\small{hult@kth.se}} \\
  \and
  Atsuto Maki \\
  Department of Computer Science\\
  KTH Royal Institute of Technology\\
  Stockholm, Sweden \\
  {\tt\small{atsuto@kth.se}} \\
}

\maketitle
\ificcvfinal\thispagestyle{empty}\fi

\tableofcontents

\clearpage
\section{Proofs}

\subsection{Proof of Theorem 1}
\noindent
First, since the components of $X$ are independent Gaussian fields with covariance kernel
\begin{align}
    k_{a,b}(\omega,\omega') = a^2 \exp\left( -\frac{\lVert \omega-\omega'\rVert_2^2}{2b^2} \right), \quad \omega, \omega'\in \mathbb{R}^n,
\end{align}
it follows that
\begin{align}
    \text{Var}((X_1(\omega),\dots,X_n(\omega))^T) = (k_{a,b}(\omega,\omega),\dots, k_{a,b}(\omega,\omega))^T = (a^2,\dots,a^2)^T,
\end{align}
and consequently that the marginal probability densities $X(\omega),\omega\in\Omega$ are given by
\begin{align}
    X(\omega) \sim p_{a^2}(x), x \in \mathbb{R}^n.
\end{align}
Now, let 
\begin{align}
    \hat{l}(\omega) = 
    \begin{cases}
        l(\omega) &\text{ if } \omega \in \Omega, \\
        0 &\text{ otherwise},
    \end{cases}
\end{align}
which means that
\begin{align}
 L(\omega) \overset{d}{=} \hat{l}(\omega + X(\omega)),
\end{align}
 where $\overset{d}{=}$ denotes equal in distribution.
This together with the law of the unconscious statistician and the symmetry $p_{a^2}(x) = p_{a^2}(-x), x\in \mathbb{R}^n$ then for any $\omega\in\Omega$ yields
\begin{align}
\mathbb{E}[L(\omega)]  &= 
\mathbb{E}[\hat{l}(\omega + X(\omega))]  \\
&= 
\int_{\mathbb{R}^n} \hat{l}(\omega+x)  p_{a^2}(x)\lambda(d x) \\
&= 
\int_{\mathbb{R}^n} \hat{l}(\omega')  p_{a^2}(\omega'-\omega)\lambda(d\omega') \\
&= 
\int_{\mathbb{R}^n} \hat{l}(\omega')  p_{a^2}(\omega-\omega')\lambda(d\omega') \\
&= 
\int_\Omega l(\omega')  p_{a^2}(\omega-\omega')\lambda(d\omega').
\end{align}
This completes the proof. \qed

\subsection{Proof of Theorem 2}
\noindent
By Theorem~1,
changing the order of integration and recalling that $\int_{\mathbb{R}^n}p_{a^2}(\omega-\omega')\lambda(d\omega) = 1$ for any $\omega'\in\mathbb{R}^n$, it follows that

\begin{align}
    \mathbb{E}\left[ \lVert L \rVert_1 \right] &=
    \mathbb{E}\left[\int_\Omega \lvert L(\omega)\rvert  \lambda(d\omega)\right] \\
    &=
    E\left[\int_\Omega  L(\omega)  \lambda(d\omega)\right] \\
    &=
    \int_\Omega \mathbb{E} [L(\omega)]  \lambda(d\omega) \\
    &=
    \int_\Omega \int_\Omega l(\omega')  p_{a^2}(\omega-\omega')\lambda(d\omega')
    \lambda(d\omega) \\
    &=
    \int_\Omega  l(\omega')  \left[\int_{\mathbb{R}^n}p_{a^2}(\omega-\omega')\lambda(d\omega) \right]\lambda(d\omega')  -
    \int_\Omega  l(\omega')  \left[\int_{\mathbb{R}^n\setminus \Omega}p_{a^2}(\omega-\omega')\lambda(d\omega) \right]\lambda(d\omega')  \\
    &=
    \int_\Omega  l(\omega') \lambda(d\omega') -\xi \\
    &=
    \int_\Omega  \lvert l(\omega') \rvert \lambda(d\omega') - \xi \\
    &=
    \lVert l \rVert_1 -  \xi.
\end{align}
\noindent
This completes the proof. \qed

\subsection{Proof of Theorem 3} 
\noindent
Independently scattered measures are introduced more generally for $\alpha$-stable distributions in \cite[Section~3.3]{samoradnitsky1994stable}.
The Gaussian case used in this work is simply the special case when $\alpha=2$.
Let $W$ be an independently scattered Gaussian measure on $\mathbb{R}^n$, with control measure $\lambda$.
That is, with $\mathcal{B}_0$ being the Lebesgue measurable sets with finite measure, for any finite collection, $A_1, \dots, A_k$ of disjoint sets in $\mathcal{B}_0$, the random variables $W(A_1), \dots, W(A_k)$ are independent, and $W(A_i)$ has centered Gaussian distribution with variance $\lambda(A_i)$.
For $f \in L^2(\mathbb{R}^n)$ the stochastic integral $I(f) = \int_{\mathbb{R}^n} f(\omega) W(d\omega)$ is well defined with centered Gaussian distribution with variance $\|f\|^2_{L^2}$. In fact, $\{I(f), f \in L^2(\mathbb{R}^n)\}$ is a Gaussian process indexed by $L^2(\mathbb{R}^n)$.
In particular, for a given $f \in L^2(\mathbb{R}^n)$, the process $Y = \{Y(\omega), \omega \in \mathbb{R}^n\}$ given by
\begin{align}
    Y(\omega) = \int_{\mathbb{R}^n} f(\omega-\omega') W(d\omega'), 
\end{align}
is a centered Gaussian process with covariance kernel given by
\begin{align}
    k(\omega, \omega') = \int_{\mathbb{R}^n} f(\omega-u)f(\omega'-u) \lambda(du).
\end{align}
For the squared exponential kernel
\begin{align}
    k(\omega, \omega') = a^2 \exp\left(-\frac{\|\omega - \omega'\|^2}{2b^2} \right)
\end{align}
we can identify $f$ as 
\begin{align}
    f(\omega) =  \frac{a}{(\pi b^2/2)^{n/4}} \exp\left(-\frac{\lVert \omega\rVert^2}{b^2}\right).
\end{align}
Indeed, for any $\omega, \omega'\in \mathbb{R}^n$ it follows, after a completion of the square, that 
\begin{align}
    k(\omega, \omega') &= \int_{\mathbb{R}^n} f(\omega-u)f(\omega'-u) \lambda(du) \\ 
    &= \frac{a^2}{(\pi b^2/2)^{n/2}}  \int_{\mathbb{R}^n} 
    \exp\left(-\frac{\|\omega - u\|_2^2}{b^2}\right)\exp\left(-\frac{\|\omega' - u\|_2^2}{b^2}\right) \lambda(du) \\
    &= \frac{a^2}{(\pi b^2/2)^{n/2}} 
    \exp\left(-\frac{\|\omega - \omega'\|_2^2}{2b^2}\right) 
    \int_{\mathbb{R}^d} 
    \exp\left(-\frac{\|u-\frac{\omega + \omega'}{2}\|_2^2}{2(b/2)^2}\right) \lambda(du) \\  
    &= \frac{a^2(2\pi (b/2)^2)^{d/2} }{(\pi b^2/2)^{d/2}} 
    \exp\left(-\frac{\|\omega - \omega'\|_2^2}{2b^2}\right) 
     \\
    &= a^2
    \exp\left(-\frac{\|\omega - \omega'\|_2^2}{2b^2}\right) 
\end{align}
Now consider the isotropic normal density in dimension $d$ with variance $b^2/2$
\begin{align}
    p_{b^2/2}(\omega) = \frac{1}{(\pi b^2)^{d/2}} \exp\left\{ - \frac{\lVert \omega \rVert_2^2}{b^2} \right\}
\end{align}
and note that it can be used to rewrite $f$ as follows
\begin{align}
    f(\omega) 
    = a (2\pi b^2)^{n/4}p_{b^2/2}(\omega).
\end{align}
Consequently, it follows that 
\begin{align}
    Y(\omega) =  a (2\pi b^2)^{d/4} \int_{\mathbb{R}^n} p_{b^2/2}(\omega-\omega') W(dw').
\end{align}
Now, let $W_1,\dots,W_n$ be independent copies of $W$, then it follows that
\begin{align}
    X(\omega) \overset{d}{=}  
    \begin{pmatrix}
    a (2\pi b^2)^{n/4} \int_{\mathbb{R}^n} p_{b^2/2}(\omega-\omega') W_1(dw') \\
    \vdots \\
    a (2\pi b^2)^{n/4} \int_{\mathbb{R}^n} p_{b^2/2}(\omega-\omega') W_n(dw')
    \end{pmatrix},
\end{align}
where $\overset{d}{=}$ denotes equal in distribution.
\noindent
This completes the proof. \qed

\section{Experiments}
\noindent
The experimental code is composed of two parts.
The first part is for extracting the data.
The second part is for training the model.
Most of the code is straight forward and similar to what would be found in any standard implementation of a UNet trained with either cross-entropy or soft-Dice.
There are however three methods based on the theory described in this paper that needs clarification.

\subsection{Computing the marginals}
\begin{lstlisting}[breaklines=true, language=Python,frame=single, label=listing:marginals, captionpos=b, caption=
Python code for computing the marginals associated with a noisy segmentation that is formed by the noise free segmentation $l$ and the noise strength parameter $a$. The code is a direct implementation of the Theorem~1.]
import numpy as np
import scipy.ndimage

def get_noisy_marginals(l,a):
    scaled_a = np.array(a)*np.array(l.shape)
    noisy_marginals = scipy.ndimage.gaussian_filter(m,scaled_a,mode='constant')

    return noisy_marginals
\end{lstlisting}

\noindent
The first method is for computing the marginals associated with a particular noisy segmentation and is based on Theorem~1.
The code for this method is listed in Listing~\ref{listing:marginals}.
The input is composed of: $l$ a discretized version of the noise-free reference segmentation represented as a multidimensional numpy array and $a$ the parameter to the noise model encoding the strength of the noise represented as a floating point number.
The output is a discretized version of the marginals represented as a multidimensional numpy array.

\subsection{Sampling noisy segmentations}

\begin{lstlisting}[breaklines=true,language=Python,frame=single,label=listing:sample,captionpos=b,caption=Python code for computing a random sample associated with a noisy segmentation that is formed by the noise free segmentation $l$ and the noise strength parameter $a$. The code is a direct implementation of Theorem~3.]
import numpy as np
import scipy.ndimage

def get_noisy_sample(l,a):
    b = 0.15*np.sqrt(2)
    scaled_a = np.array(a)*np.array(l.shape)
    scaled_b = np.array(b)*np.array(l.shape)
    weight = scaled_a*(2*np.pi*scaled_b**2)**(len(np.shape(m))/4)

    perb = np.array([weight*scipy.ndimage.gaussian_filter(
        np.random.normal(size=l.shape),scaled_b[i]/np.sqrt(2),mode='constant')
        for i in range(len(l.shape))])
    grid_mesh = np.meshgrid(*[range(l.shape[i]) for i in range(len(l.shape))],indexing='ij')
    noisy_sample = np.round(scipy.ndimage.map_coordinates(l,grid_mesh+perb, mode='nearest'))

    return noisy_sample
\end{lstlisting}

\noindent
The second method is for sampling noisy segmentations and is based on Theorem~3. 
The code for this method is listed in Listing~\ref{listing:sample}.
The input is composed of: $l$ a discretized version of the noise-free reference segmentation represented as a multidimensional numpy array and $a$ the parameter to the noise model encoding the strength of the noise represented as a floating point number.
The output is a discretized version of a random sample associated with the noisy segmentation and is represented as a multidimensional numpy array.
The method can be broken down into three steps.
Firstly the constant used for rescaling is computed.
Secondly, the vector of Gaussian fields is generated, which in the numerical setting is approximated by drawing i.i.d. Gaussian variables for each entry in $l$ and then processing the resulting array with a Gaussian filter.
Thirdly, the noise-free segmentation $l$ is deformed with the resulting random deformation array.

\subsection{Dice optimal segmentations}

\begin{lstlisting}[breaklines=true,language=Python,frame=single,label=listing:diceopt,captionpos=b,caption=Python code for generating a Dice optimal segmentation from a marginal function $m$.]
import numpy as np

def get_opt_dice_seg(m):
    psi = np.flip(np.sort(m.flatten()))
    d = 2*np.cumsum(psi)/(np.sum(m)+np.arange(1,len(psi)+1))
    t = np.max(d)/2
    s = 1.0*(m>=t)
    return s
\end{lstlisting}

\noindent
The third method is for computing the optimal segmentation with respect to Dice.
The code for this method is listed in Listing~\ref{listing:diceopt} and is taken from~\cite{nordstrom2022image}.
It is an efficient variation of a method proposed in binary classification~\cite{lipton2014optimal}.
The input is composed of: $m$ a discretized version of the the marginal function represented as a multidimensional numpy array.
The output is a discretized version of the optimal segmentation with respect to Dice represented as a multidimensional numpy array.
The idea is to sort the voxels in $m$ from largest to smallest, and then compute the Dice score associated with the segmentations that are formed by taking the first set of voxels associated with this sorted list.
This is done for every possible number of voxels.
The maximal score is used to form a threshold which is used to formulate the final segmentation.

{\small
\bibliographystyle{ieee_fullname}
\bibliography{bibliography}
}